%% file: main.tex
\documentclass{article}

\usepackage[preprint]{neurips_2026}

\usepackage[utf8]{inputenc}
\usepackage[T1]{fontenc}
\usepackage{microtype}
\usepackage{amsmath,amssymb,amsfonts}
\usepackage{amsthm}
\usepackage{bm}
\usepackage{booktabs}
\usepackage{multirow}
\usepackage{graphicx}
\usepackage{comment}
\usepackage{xcolor}
\usepackage{hyperref}
\usepackage{url}
\usepackage{algorithm2e}
\usepackage{enumitem}
\usepackage{natbib}
\usepackage{subcaption}
\usepackage{wrapfig}
\usepackage{colortbl}
\usepackage{float}

\newcommand{\enc}{\mathrm{enc}}
\newcommand{\pred}{\mathrm{pred}}
\newcommand{\idm}{\mathrm{idm}}
\newcommand{\gcidm}{\mathrm{gc\text{-}idm}}
\newcommand{\vo}{\bm{o}}
\newcommand{\vz}{\bm{z}}
\newcommand{\va}{\bm{a}}
\newcommand{\vc}{\bm{c}}
\newcommand{\vh}{\bm{h}}
\newcommand{\gL}{\mathcal{L}}
\newcommand{\gD}{\mathcal{D}}
\newcommand{\mZ}{\bm{Z}}
\newcommand{\R}{\mathbb{R}}
\newcommand{\E}{\mathbb{E}}

\theoremstyle{plain}

\newtheorem{proposition}{Proposition}

\theoremstyle{definition}
\newtheorem{assumption}{Assumption}

\theoremstyle{remark}

\definecolor{bestcolor}{HTML}{E8F5E9}

\hypersetup{
    colorlinks=true,
    linkcolor=blue!60!black,
    citecolor=blue!60!black,
    urlcolor=blue!60!black,
}

\AtEndEnvironment{table}{\vspace{1mm}}
\AtEndEnvironment{table*}{\vspace{1mm}}

\title{Latent Geometry Beyond Search: \\Amortizing Planning in World Models}

\author{%
  Hoang Nguyen\thanks{Equal Contribution}\quad Xiaohao Xu${}^{*}$\thanks{Project Lead}\quad Xiaonan Huang \\ Department of Robotics, University of Michigan, Ann Arbor
}

\begin{document}
\maketitle


\begin{abstract}
Modern vision-based world models can represent observations as compact yet expressive latent manifolds, but fast goal-oriented planning in these spaces remains challenging. This raises a central question: when does a learned representation simplify control, rather than merely enabling prediction? We study this question in a pretrained LeWorldModel, whose latent geometry is regularized for smoothness and uniformity.
Our key insight is that, under such geometry, planning can be amortized into a latent inverse-dynamics mapping instead of requiring online search. We therefore replace iterative planning with a lightweight Goal-Conditioned Inverse Dynamics Model (GC-IDM) that maps the current latent state, goal latent state, and remaining horizon directly to the next action.
Empirically, across four benchmark environments spanning navigation, contact-rich manipulation, and continuous control, our controller matches or exceeds CEM in seven of eight environment–protocol settings while reducing per-decision cost by $100$–$130\times$. A broader sweep over test-time planners (CEM, MPPI, iCEM, and gradient-based methods) shows that this result is not specific to a particular optimizer. These findings suggest that much of the structure recovered by test-time planning is already locally encoded in the latent representation. More broadly, our results indicate that sufficiently structured latent spaces can shift part of the planning burden from online optimization to learned inference.
\end{abstract}

\section{Introduction}
\label{sec:intro}
\textcolor{black}{The utility of world models in control ultimately depends on their ability to facilitate efficient decision-making.} Predicting future observations or latent states is useful, but it does not by itself solve the control problem: an agent must still convert those predictions into actions. In many current systems, this final step remains \textcolor{black}{computationally} expensive. Learned representations make forecasting accurate and fast, yet action selection still relies on substantial online optimization. As a result, there is often a gap between learning a good model of the world and using that model to act efficiently.

\textcolor{black}{The LeWorldModel} (LeWM)~\citep{lewm2026} is an attractive starting point because it is already efficient as a world model: with ${\sim}15$M parameters trainable on a single GPU in a few hours, it plans up to $48\times$ faster than foundation-model-based world models while remaining competitive across diverse control tasks. Yet even in LeWM, control is still implemented through CEM, which requires $9{,}000$ candidate rollouts ($45{,}000$ predictor forward passes) for every planning step. This means that the dominant computational cost no longer comes from modeling the dynamics; it comes from searching over candidate action sequences. We refer to this mismatch between cheap prediction and expensive decision-making as the \emph{planning tax}.

This raises a basic question: is \textcolor{black}{this} \emph{tax} intrinsic, or is it \textcolor{black}{primarily} an artifact of how control is formulated? If the latent space is poorly organized, then search may indeed remain necessary, because the mapping from desired future states to actions is unstable and nonlocal. But if the latent space is smooth and action-sensitive, properties that regularizers like SIGReg are designed to encourage, then action selection may admit a simpler description. In that case, planning can be viewed not as a generic combinatorial search problem, but as a local inverse problem induced by representation regularity.
We study this perspective in the setting of a pretrained JEPA world model. In this paper, we replace online search with a small goal-conditioned inverse dynamics model \textcolor{black}{(GC-IDM)} trained on frozen LeWM latents. The model takes the current latent state, the goal latent state, and the remaining horizon, and predicts the next action directly. \textcolor{black}{Our} central question is \textbf{\textit{whether inverse dynamics \textcolor{black}{serves as} a sufficiently faithful abstraction of control for well-regularized latent representations}}.

Across four benchmark environments, \textcolor{black}{the proposed GC-IDM} matches or exceeds CEM in seven of \textcolor{black}{the} eight environment--protocol cells while reducing \textcolor{black}{per-decision} planning cost by $100$--$130\times$. The same conclusion holds under a strict episode-level holdout, under a broad CEM compute sweep across a $500\times$ range, and across \textcolor{black}{a} broader family of test-time planners including CEM~\citep{rubinstein1999cem},  iCEM~\citep{pinneri2021icem}, MPPI~\citep{williams2017mppi}, \textcolor{black}{and} gradient-based trajectory optimization~\citep{hansen2022tdmpc,hansen2024tdmpc2}. Together, these results suggest that, in this regime, the planner is recovering structure that the representation has already made locally accessible.
On the empirical side, we confirm that a small inverse dynamics model trained by supervised regression on frozen world model latents can replace test-time search \textcolor{black}{while achieving comparable or superior success rates} across a broad range of benchmark conditions \textcolor{black}{and maintaining} high planning efficiency.

\paragraph{Our main contributions are:} \textbf{1) }We propose GC-IDM, a learned planning method for JEPA world models that recasts control as an inverse problem in latent space, replacing iterative test-time search with a single forward pass per environment step.
\textbf{2) }We empirically demonstrate that a \textcolor{black}{lightweight} goal-conditioned inverse model can outperform the broader family of test-time planners (including CEM, MPPI, iCEM, and gradient-based methods) on the LeWM benchmarks, achieve strong performance relative to prior baselines, and reduce planning cost by $100$--$130\times$.
\textbf{3) }Our experiments confirm that learning latent inverse dynamics on well-regularized latent representations enables fast\textcolor{black}{, effective} closed-loop planning.


\section{Related Work}
\label{sec:related}
\paragraph{Latent world models and planning over world models.}
Our work sits inside a long tradition that treats a world model as something the agent plans \emph{with}, and in that setting\textcolor{black}{,} the choice of planner is often where most of the inference cost resides. Latent world models learn environment dynamics in a compressed embedding space, allowing the agent to ``imagine'' future states and evaluate candidate behaviors through rollouts~\citep{ha2018world,hafner2019dream,hafner2020dreamer,hafner2023dreamerv3}. Test-time planning is then typically implemented with model-predictive control using CEM~\citep{rubinstein1999cem,pinneri2021icem}, MPPI~\citep{williams2017mppi}, or gradient-based trajectory optimization~\citep{hansen2022tdmpc,hansen2024tdmpc2}. Prior JEPA-style world models either rely on pretrained encoders to stabilize the representation~\citep{zhou2024dinowm,vjepa2} or use more elaborate multi-term objectives with heuristic regularization~\citep{sobal2022slowfeatures,sobal2025pldm}. LeWM~\citep{lewm2026} addresses both issues with a simple two-term objective based on latent future-embedding prediction and SIGReg regularization~\citep{balestriero2025lejepa}, and it is the backbone we build \textcolor{black}{upon}. Our work is orthogonal to this line of \textcolor{black}{research}: we do not propose a new world model or a new optimizer, but instead ask whether the optimizer can be removed because the learned representation \textcolor{black}{has effectively} absorbed the computation it performs.

\paragraph{Inverse dynamics for control.}
The idea that inverse dynamics can support or replace explicit planning has been explored in several regimes, but typically as a proposal mechanism, an auxiliary prediction head, or a component of imitation learning rather than as the \textcolor{black}{primary} planner itself inside a JEPA latent world model. Inverse-dynamics models predict the action responsible for a state transition and have been widely used in model-based reinforcement learning and representation learning~\citep{agrawal2016curiosity,pathak2017curiosity,baker2022idm}. GLAMOR~\citep{paster2021glamor} is the closest precursor: it learns a recurrent inverse-dynamics model from pixels that predicts entire action sequences conditioned on a goal state, combining the IDM with a learned action prior for planning. Our approach differs in three respects: (i) we predict a single action per step and replan from a fresh observation, rather than committing to a multi-step action sequence; (ii) our IDM is a simple feedforward MLP rather than a recurrent network; and (iii) we operate in the latent space of a frozen pretrained JEPA world model rather than learning the representation end-to-end. Predictive inverse-dynamics methods such as Seer~\citep{tian2024seer} and Latent Diffusion Planning~\citep{xie2025ldp} first forecast a future state and then decode an action from that prediction. By contrast, our model conditions directly on the terminal goal embedding together with the remaining horizon and relies on closed-loop re-encoding at every step, avoiding an explicit state-prediction stage inside the controller. Mimic-video~\citep{pai2025mimicvideo} is especially relevant in showing that a pretrained visual backbone paired with a lightweight inverse model can yield strong control performance. Recent theory also helps explain when this strategy should work: \citet{pidm_theory2026} analyze a bias--variance tradeoff in future-state-conditioned inverse dynamics, and \citet{morin2026idmsample} argue that ground-truth inverse dynamics often lie in a lower-complexity hypothesis class than the expert policy. These perspectives are consistent with our findings that a simple inverse model can be effective when paired with a well-structured latent representation. Closely related, \citet{luo2025grounding} also ground a pretrained visual world model into executable actions via an inverse-dynamics-style decoder, but train it action-label-free through self-supervised goal-conditioned exploration; we instead train offline on the LeWM dataset's action labels and use the IDM to replace the CEM optimizer entirely.

\section{\textcolor{black}{Preliminaries}}
\label{sec:background}
\paragraph{Problem formulation.}
We consider finite-horizon goal-conditioned control from observations. At time $t$, the agent observes $\vo_t \in \mathcal{O}$, aims to reach a goal observation $\vo_g \in \mathcal{O}$ within a remaining horizon $h_t \in \{1,\dots,H\}$, and selects an action $\va_t \in \mathcal{A}$. An encoder $\enc_\theta : \mathcal{O} \to \R^d$ maps observations to latent states $\vz_t = \enc_\theta(\vo_t)$ and $\vz_g = \enc_\theta(\vo_g)$. The latent-planning problem is to choose actions that minimize goal mismatch in latent space:
\begin{equation}
    \pi^\star(\vo_t, \vo_g, h_t)
    \in
    \arg\min_{\pi}
    \; \E\!\left[\sum_{k=0}^{h_t-1} c\bigl(\vz_{t+k}, \vz_g\bigr)\right],
    \label{eq:problem_formulation}
\end{equation}
where $c : \R^d \times \R^d \to \R_{\ge 0}$ is a goal-matching cost, typically induced by Euclidean distance in latent space. Search-based planners approximately solve Equation~(\ref{eq:problem_formulation}) online by repeatedly rolling out a learned latent dynamics model. When the latent geometry is sufficiently regular, the optimal first action may instead be recovered directly by a learned inverse map from $(\vz_t, \vz_g, h_t)$.

\noindent\textbf{Latent world model.} We consider a latent world model with an encoder $\enc_\theta$ that maps observations to latent states and a predictor $\pred_\phi$ that models one-step latent dynamics under actions:
\begin{equation}
    \vz_t = \enc_\theta(\vo_t), \qquad \hat{\vz}_{t+1} = \pred_\phi(\vz_t, \va_t).
    \label{eq:lewm}
\end{equation}
The encoder and predictor are jointly optimized using
\begin{equation}
    \gL_{\text{LeWM}} \triangleq \gL_{\text{pred}} + \lambda \, \text{SIGReg}(\mZ),
    \label{eq:lewm_loss}
\end{equation}
Here $\gL_{\text{pred}} = \|\hat{\vz}_{t+1} - \vz_{t+1}\|_2^2$ is the one-step latent prediction loss, $\hat{\vz}_{t+1}$ is the predictor output, $\vz_{t+1}$ is the encoded next observation, and $\mZ$ denotes the batch of latent embeddings on which the regularizer is evaluated. The term $\text{SIGReg}(\mZ)$ is the isotropy-inducing regularizer of~\citet{balestriero2025lejepa}, and $\lambda > 0$ is its scalar weighting coefficient relative to the prediction term. Thus, $\lambda$ controls the tradeoff between predictive accuracy and regularity of the latent geometry. The geometric consequence is that the learned latent space is smooth, regularized toward well-distributed embeddings, and directly optimized for predictive dynamics.
\paragraph{Search-based planning in latent space.}
Given a current latent state $\vz_t$ and a goal latent state $\vz_g$, a standard planner chooses an action sequence by minimizing a terminal latent cost through repeated rollouts of the predictor:
\begin{equation}
    \va_{t:t+H-1}^{\star} \in \arg\min_{\va_{t:t+H-1}} d\!\left(\hat{\vz}_{t+H}(\va_{t:t+H-1}), \vz_g\right).
    \label{eq:latent_planning_problem}
\end{equation}
Here $d(\cdot,\cdot)$ is the latent goal-distance metric, $H$ is the planning horizon, and $\hat{\vz}_{t+H}(\va_{t:t+H-1})$ denotes the latent state predicted after rolling out the action sequence $\va_{t:t+H-1}$ through the learned dynamics model. In practice this optimization is usually approximated with sampling-based MPC such as CEM~\citep{rubinstein1999cem}. With LeWM's default configuration ($300$ samples, $30$ refinement iterations, planning horizon $5$), this comes to $9{,}000$ candidate rollouts and $45{,}000$ predictor forward passes per planning step. When the latent space is sufficiently regular, the optimizer in Equation~(\ref{eq:latent_planning_problem}) can be replaced by a learned inverse map.
\section{Method: Goal-Conditioned Inverse Dynamics Model (GC-IDM)}
\label{sec:method}
\begin{figure}[t]
\centering
\includegraphics[width=\linewidth]{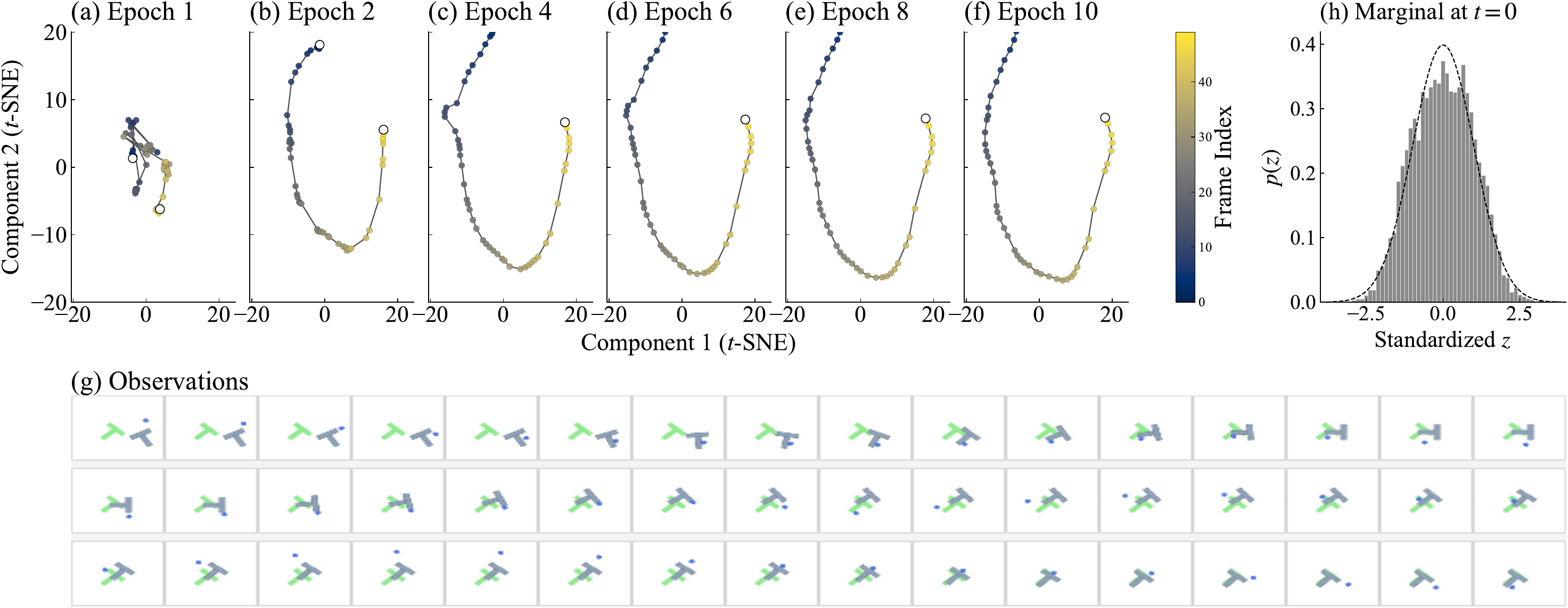}
\caption{\textbf{Evolution of Push-T latent geometry across training.} Panels (a)--(f) show two-dimensional t-SNE embeddings of latent states from Push-T sequence at epochs 1, 2, 4, 6, 8, and 10, with points colored by frame index. Panel (g) shows subsampled observation frames from the same sequence. Panel (h) shows the standardized marginal latent distribution at $t{=}0$ for epoch 10, together with a Gaussian reference curve. Across training, the embedding evolves from a diffuse cloud into a smooth ordered path aligned with task progress, providing qualitative support for the claim that the learned representation is structured enough for local inverse recovery.}
\label{fig:qualitative_latents}
\end{figure}
\begin{figure}[t]
    \centering
    
    \resizebox{\textwidth}{!}{%
        \includegraphics[height=5cm]{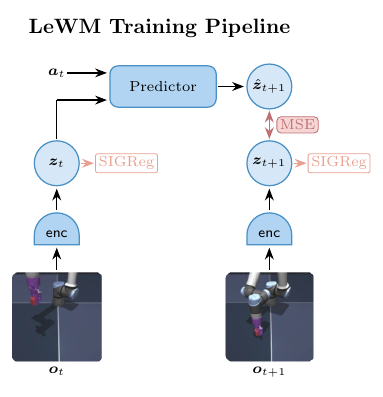}%
        \includegraphics[height=5cm]{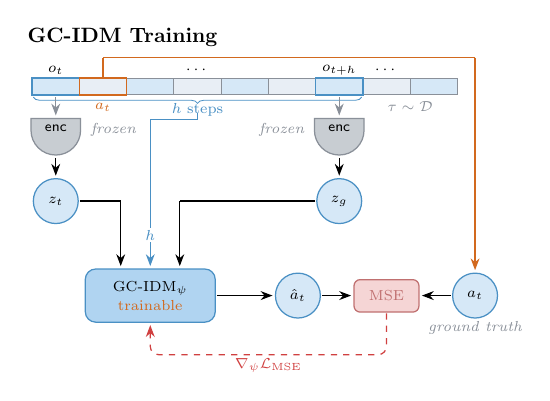}%
        \includegraphics[height=5cm]{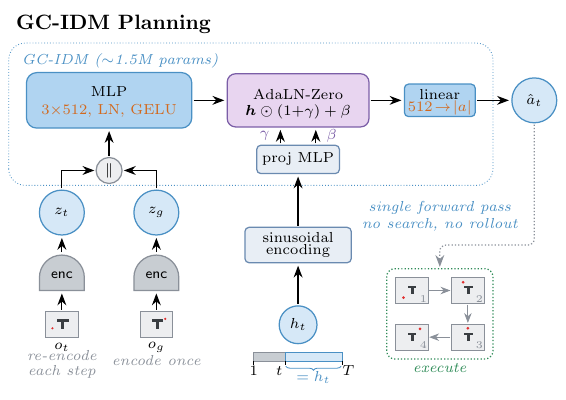}%
    }
    
    \caption{\textbf{Pipeline overview.} \emph{Left:} World model encoder training process, which follows LeWM \citep{lewm2026}. \emph{Center:} Goal-conditioned inverse dynamics model (GC-IDM) training. From a trajectory $\tau \sim \mathcal{D}$, a tuple $(\vz_t, \vz_g, h, \va_t)$ is sampled at random horizon $h \in [1, H_{\max}]$ using frozen LeWM embeddings; the IDM is trained by MSE regression with gradients flowing only into the inverse dynamics module, i.e.,  $\text{GC-IDM}_\psi$. \emph{Right:} GC-IDM planning. The current observation $o_t$  is re-encoded at every step; the goal embedding $z_g$ is encoded once and cached. The concatenated pair $\vz_t \| \vz_g$ passes through a $3$-layer MLP modulated by the remaining horizon $h_t$ via AdaLN-Zero, producing estimated action $\hat{\va}_t$ in a single forward inference pass with no search or rollout.}
    \label{fig:three_images}
\end{figure}
\subsection{Overview}
\label{sec:gcidm_overview}
\noindent\textbf{Motivation.} Our motivation is to remove the planning tax without \textcolor{black}{altering} the pretrained world model. A useful failure case clarifies \textcolor{black}{our} design: our first attempt (Appendix~\ref{app:pairwise}) was a \emph{pairwise} IDM that decoded actions from consecutive latent pairs $(\vz_t, \vz_{t+1}) \to \va_t$ and then planned by linearly interpolating between start and goal embeddings. That model achieved near-perfect oracle reconstruction ($R^2 = 0.993$) but poor planning. This supports the idea that bottleneck is not decoding actions from well-posed local transitions; it is constructing a valid latent path between distant states. 

\noindent\textbf{Latent geometry intuition.} If the pretrained latent geometry is already smooth and task-aligned, then the controller does not need to invent an entire trajectory between distant states; it only needs to recover the locally appropriate action that moves the current state toward the goal under a finite budget. Figure~\ref{fig:qualitative_latents} provides qualitative evidence for this picture on Push-T by showing how the t-SNE embedding of one sequence evolves across training for a pretrained LeWM, together with subsampled observations and a marginal latent histogram. Across epochs, the latent states organize from a diffuse cloud into a smooth ordered path aligned with task evolution, suggesting that nearby latent displacements correspond to locally coherent behavioral changes. That is, it supports the motivating hypothesis that sufficiently regular latent geometry can make local inverse recovery easy to amortize.

\noindent\textbf{Controller.} Our replacement for search is, therefore, a goal-conditioned inverse dynamics operator that acts directly on the current latent state, the goal latent state, and the remaining horizon. Given observations $(\vo_t, \vo_\text{goal})$, we encode them with the frozen encoder and predict the next action by
\begin{equation}
    \vz_t = \enc_\theta(\vo_t), \quad \vz_\text{goal} = \enc_\theta(\vo_\text{goal}), \quad \hat{\va}_t = \gcidm_\psi\!\bigl(\vz_t,\, \vz_\text{goal},\, h_t\bigr).
    \label{eq:gcidm}
\end{equation}
Equation~(\ref{eq:gcidm}) is the full inference-time model. Relative to pairwise interpolation, it avoids imagined latent trajectories and conditions only on real encoder outputs from actual observations. The resulting policy is still goal-conditioned, but it shifts the problem from online search over action sequences to offline amortization of the local inverse map induced by the latent world model. 
\subsection{ Model Architecture and Details}
\label{sec:gcidm_arch}
\RestyleAlgo{ruled}
\begin{algorithm}[t]
\small
\DontPrintSemicolon
\SetAlgoLined
\KwIn{Environment $\mathcal{E}$, encoder $\enc_\theta$, inverse policy $\gcidm_\psi$, goal observation $\vo_g$, horizon budget $T$}
\KwOut{Closed-loop action sequence $(\va_1,\ldots,\va_\tau)$, where $\tau \le T$}
$\vz_g \gets \enc_\theta(\vo_g)$\;
reset or initialize $\mathcal{E}$ and observe initial observation $\vo_1$\;
\For{$t \gets 1$ \KwTo $T$}{
    $\vz_t \gets \enc_\theta(\vo_t)$\;
    $h_t \gets T - t + 1$\;
    $\va_t \gets \gcidm_\psi(\vz_t, \vz_g, h_t)$\;
    apply $\va_t$ to $\mathcal{E}$ and observe next observation $\vo_{t+1}$\;
    \If{goal reached or episode terminated}{
        \Return $(\va_1,\ldots,\va_t)$\;
    }
}
\Return $(\va_1,\ldots,\va_T)$\;
\caption{Closed-loop goal-reaching control induced by GC-IDM}
\label{alg:gcidm}
\end{algorithm}
We \textcolor{black}{purposefully} instantiate Equation~(\ref{eq:gcidm}) with a \textcolor{black}{lightweight} neural network (Figure~\ref{fig:three_images}) so that any gain over CEM is attributable to latent geometry and control structure rather than raw function capacity. The backbone is a 3-layer MLP with hidden dimension $512$, LayerNorm, GELU activation, and $10\%$ dropout, with the two embeddings concatenated as $\vz_t \,\|\, \vz_\text{goal}$ at the input. A final linear head maps the modulated representation to the action space. Total parameters: ${\sim}1.5$M, roughly $10\%$ of the LeWM backbone and orders of magnitude smaller than the ${\sim}10$M-parameter predictor that CEM invokes $45{,}000$ times per plan call. Given a pretrained LeWM checkpoint, GC-IDM can be trained from scratch in approximately 20 minutes per environment on a single GPU, making it a lightweight extension of an already efficient world model.
The remaining-horizon variable modulates the inverse map rather than being concatenated as an undifferentiated observation feature. Concretely, the remaining-step count is first normalized as $h_t = \min(\text{steps\_remaining}, H_\text{max}) / H_\text{max} \in [0,1]$, sinusoidally encoded ($64$ dimensions), and passed through a 2-layer MLP to produce an embedding $\vc$. Two separate linear projections $\gamma(\vc), \beta(\vc)$ then modulate the backbone output $\vh$ via zero-initialized AdaLN-Zero modulation~\citep{peebles2023dit}:
\begin{equation}
    \mathrm{AdaLN}_0(\vh, \vc) \;=\; \vh \,\odot\, \bigl(\mathbf{1} + \gamma(\vc)\bigr) \;+\; \beta(\vc),
    \label{eq:adaln}
\end{equation}
applied immediately before the action head. The two projection layers $\gamma(\cdot), \beta(\cdot)$ have their weights {and} biases zero-initialized, so at the start of training $\gamma{=}\beta{=}\textbf{0}$, ensuring the horizon signal is introduced only as the regression loss demands it.
\paragraph{Training on frozen embeddings.} Training uses the same offline demonstration dataset that trained LeWM; no additional environment interaction is required. Each training example is induced from a trajectory by sampling $(t,h)$, with $h$ drawn uniformly from $[1, H_\text{max}]$, and forming the triple $(\vz_t, \vz_{t+h}, \va_t)$. The loss function is
\begin{equation}
    \gL_{\text{gc-idm}}(\psi) = \E_{(t, h) \sim \gD}\left[\,\bigl\|\gcidm_\psi(\vz_t, \vz_{t+h}, h) - \va_t\bigr\|_2^2\,\right].
    \label{eq:gcidm_loss}
\end{equation}

\paragraph{Closed-loop control at test time.} At test time, the trained inverse map replaces the entire iterative optimization loop. Control is implemented as a receding-horizon inverse policy: given a goal observation $\vo_g$ and a budget $T$, the controller encodes the goal once, then repeatedly re-encodes the current observation, evaluates the inverse map at the current latent state and remaining horizon, and applies the resulting action to the environment. The remaining horizon is computed as $h_t = T - t + 1$ and clamped to $H_\text{max}$ (matching the training distribution); in all experiments $H_\text{max} = T = 50$. There is no inner optimization loop and no open-loop commitment to a predicted latent trajectory. Algorithm~\ref{alg:gcidm} states the control law formally.
Algorithm~\ref{alg:gcidm} is a closed-loop controller: after each action, it re-encodes the actual observation and recomputes the next move from the true current state. In our implementation, the added computation is a ${\sim}1.5$M-parameter MLP on top of encoder calls already required by LeWM. Intuitively, the latent goal distance $\|\vz_t - \vz_g\|$ acts as a potential field, and GC-IDM approximately follows its gradient at each step; per-step replanning accumulates these directionally correct predictions into successful trajectories.

\section{Experiments}
\label{sec:experiments}

\subsection{Experimental Setup}
\label{sec:setup}

\textcolor{black}{Consistent with} the LeWM~\citep{lewm2026} benchmark, each evaluation environment \textcolor{black}{requires} goal-conditioned control from image observations, \textcolor{black}{though} they differ in action dimensions and contact structures: Two-Room is nearly deterministic navigation, Push-T requires extended contact-rich manipulation, OGBench-Cube stresses 3D object manipulation, and Reacher is continuous reaching. 

For \textcolor{black}{each} environment, we train GC-IDM on frozen LeWM embeddings and evaluate \textcolor{black}{it} against the same pretrained LeWM backbone \textcolor{black}{using its original default} CEM planner configuration. We report \textcolor{black}{results across} two evaluation protocols ($n{=}50$ and $n{=}200$), a strict episode-level holdout, and \textcolor{black}{present supplementary} compute and ablation studies in the appendix.

\begin{table}[t]
\centering
\caption{\textbf{Main results.} GC-IDM vs.~CEM on the four LeWM benchmark environments. Both methods use the frozen LeWM checkpoint from~\citet{lewm2026} and the same solver configuration. We experiment each environment at two sample sizes: $n{=}200$ is our headline reporting protocol, chosen for lower variance; $n{=}50$ is the LeWM paper's own protocol. Both GC-IDM and CEM are mean $\pm$ standard deviation across three seeds. Our method's columns are underlined throughout for visual tracking; highlighted cells mark the winning method per row. Two speed up measures are recorded: \emph{wall / ep.}\ is the end-to-end per-episode wall-clock ratio; \emph{ms / plan} is the average wall-clock of a single \texttt{get\_action} call (one per environment step for both methods). Because CEM buffers $25$ raw actions per solver call, its average includes both expensive solver steps and cheap buffer returns; the reported $100$--$130\times$ speedup is therefore already amortized over CEM's commit window.} \vspace{1mm}
\label{tab:headline}
\footnotesize
\setlength{\tabcolsep}{3pt}
\begin{tabular}{@{}llcccccccccc@{}}
\toprule
& & \multicolumn{2}{c}{Success Rate (\%)} & \multicolumn{2}{c}{ms / ep.} & \multicolumn{2}{c}{ms / plan} & \multicolumn{2}{c}{Speedup} & \multicolumn{2}{c}{Planner cost} \\
\cmidrule(lr){3-4} \cmidrule(lr){5-6} \cmidrule(lr){7-8} \cmidrule(lr){9-10} \cmidrule(lr){11-12}
Env & $n$ & Ours & CEM & Ours & CEM & Ours & CEM & wall & plan & Ours & CEM \\
\midrule
\multirow{2}{*}{Two-Room}
    &  50 & \cellcolor{bestcolor}\underline{\textbf{100.0$\pm$0.0}} & 82.0$\pm$2.0 & \underline{\textbf{260}}  & 10486 & \underline{\textbf{86}}  & 10317 & $40\times$  & $120\times$ & \multirow{8}{*}{\shortstack{1.5M MLP\\[2pt]0.71\,ms /fwd\\[2pt]0 pred.\ calls}} & \multirow{8}{*}{\shortstack{10.8M model\\[2pt]45K calls\\/ plan}} \\
    & 200 & \cellcolor{bestcolor}\underline{\textbf{100.0$\pm$0.0}} & 84.0$\pm$2.8 & \underline{\textbf{276}}  & 10502 & \underline{\textbf{398}} & 41318 & $38\times$  & $104\times$ & & \\
\cmidrule(lr){1-10}
\multirow{2}{*}{Push-T}
    &  50 & \underline{84.7$\pm$5.0} & \cellcolor{bestcolor}\textbf{89.3$\pm$6.4} & \underline{\textbf{305}}  & 10558 & \underline{\textbf{86}}  & 10340 & $35\times$  & $121\times$ & & \\
    & 200 & \cellcolor{bestcolor}\underline{\textbf{84.2$\pm$2.8}} & 82.5$\pm$1.3 & \underline{\textbf{322}}  & 10778 & \underline{\textbf{397}} & 42232 & $33\times$  & $106\times$ & & \\
\cmidrule(lr){1-10}
\multirow{2}{*}{OGB-Cube}
    &  50 & \cellcolor{bestcolor}\underline{\textbf{99.3$\pm$1.2}}  & 73.3$\pm$9.0 & \underline{\textbf{8587}} & 19491 & \underline{\textbf{84}}  & 10799 & $2.3\times$ & $129\times$ & & \\
    & 200 & \cellcolor{bestcolor}\underline{\textbf{98.7$\pm$0.6}}  & 67.0$\pm$2.1 & \underline{\textbf{9172}} & 20057 & \underline{\textbf{334}} & 43201 & $2.2\times$ & $130\times$ & & \\
\cmidrule(lr){1-10}
\multirow{2}{*}{Reacher}
    &  50 & \cellcolor{bestcolor}\underline{\textbf{100.0$\pm$0.0}} & 68.0$\pm$9.2 & \underline{\textbf{1365}} & 12031 & \underline{\textbf{93}}  & 10745 & $8.8\times$ & $116\times$ & & \\
    & 200 & \cellcolor{bestcolor}\underline{\textbf{99.7$\pm$0.3}}  & 70.3$\pm$4.3 & \underline{\textbf{1604}} & 12484 & \underline{\textbf{399}} & 43969 & $7.8\times$ & $110\times$ & & \\
\bottomrule
\end{tabular}\vspace{-5mm}
\end{table}

\subsection{Implementation Details}
\label{sec:exp_impl_details}

\textcolor{black}{LeWM employs} a ViT-based encoder and a transformer\textcolor{black}{-based} predictor\textcolor{black}{, whereas} GC-IDM is instantiated as a \textcolor{black}{compact} MLP with horizon conditioning. \textcolor{black}{Comprehensive details regarding} numerical hyperparameters, optimizer settings, and evaluation \textcolor{black}{configurations} are \textcolor{black}{provided} in Appendix~\ref{app:implementation}.

\paragraph{Compute accounting.}
GC-IDM incurs an additional one-time supervised fitting cost\textcolor{black}{, though it relies} on the \textcolor{black}{same} pretrained LeWM encoder and predictor shared \textcolor{black}{with CEM}. In our runs\textcolor{black}{,} this extra training stage takes approximately 20 minutes on a single GPU per environment. \textcolor{black}{Our primary comparison, therefore, contrasts} amortized offline fitting \textcolor{black}{with} repeated online search. At inference time, GC-IDM executes one encoder pass and one MLP forward pass per control step, whereas CEM repeatedly invokes the latent predictor inside its optimization loop. All main experiments are run on an NVIDIA L4 GPU for Two-Room, Push-T, and Reacher, and on an NVIDIA A100 for OGBench-Cube (which requires more GPU memory for MuJoCo rendering and physics).

\subsection{Main Results}
\label{sec:main_results}

Table~\ref{tab:headline} summarizes \textcolor{black}{this comparative analysis}. GC-IDM matches or \textcolor{black}{outperforms} CEM in seven of \textcolor{black}{the} eight environment--protocol cells while reducing \textcolor{black}{the} planning cost by $100$--$130\times$ per decision. The strongest gains occur on OGBench-Cube and Reacher, where GC-IDM is much faster and \textcolor{black}{significantly} more successful. The only exception is Push-T at $n{=}50$, where CEM is modestly better in mean success but remains more than two orders of magnitude slower in \textcolor{black}{terms of} planning time.

Two aspects of Table~\ref{tab:headline} stand out. 
First, the \textcolor{black}{advantages are} not merely computational: on OGBench-Cube and Reacher, GC-IDM is also substantially \emph{more accurate}, \textcolor{black}{challenging} the \textcolor{black}{assumption} that CEM \textcolor{black}{extracts} essential long-horizon reasoning \textcolor{black}{inaccessible} to \textcolor{black}{a local} inverse model. Second, the one environment where CEM remains competitive, Push-T, is exactly the regime in which longer contact sequences make local inverse recovery \textcolor{black}{inherently more challenging}. This pattern \textcolor{black}{supports our core hypothesis:} inverse dynamics \textcolor{black}{excels} when \textcolor{black}{the} latent geometry \textcolor{black}{ensures} well\textcolor{black}{-}conditioned \textcolor{black}{local action recovery}, \textcolor{black}{but its efficacy wanes} when the task requires more delicate multi-step contact planning.

\begin{figure}[t]
\centering
\includegraphics[width=\linewidth]{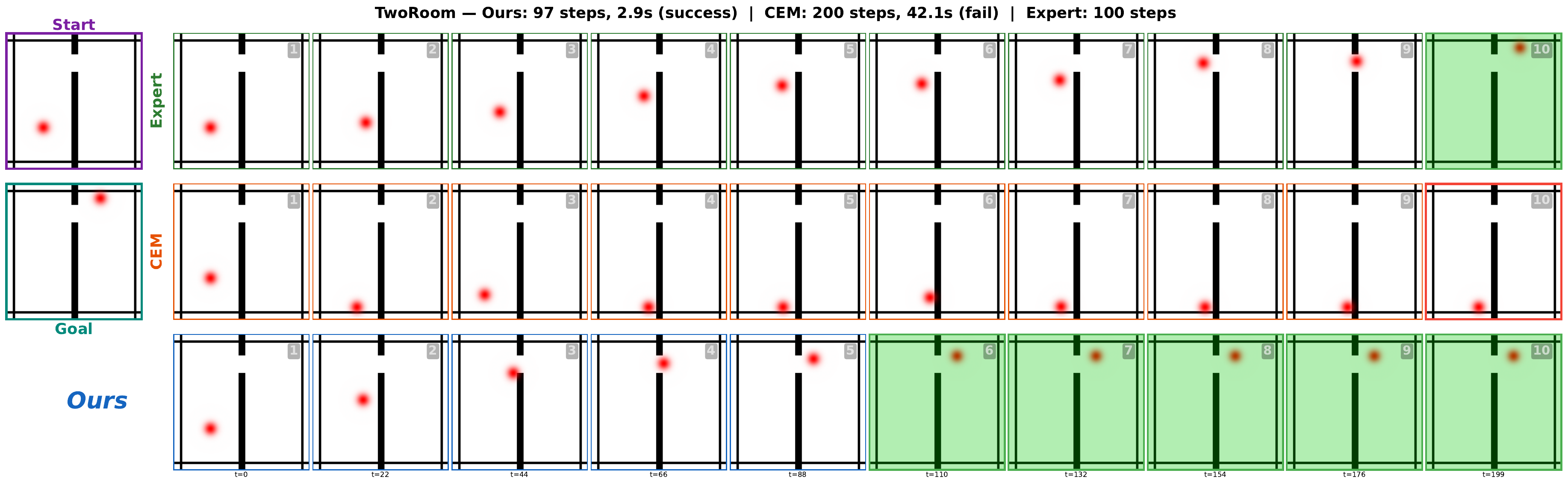}
\caption{\textbf{Matched execution rollout on Two-Room.} Expert, CEM, and GC-IDM (ours) on the same episode from identical start and goal states. The first column shows the start (top) and goal (bottom); columns 1--10 are evenly spaced frames. Green shading marks success; red borders mark failure. CEM and GC-IDM share the same time axis; the expert row uses the dataset's own time axis. GC-IDM reaches the goal in fewer steps with visibly smoother motion. On this environment, GC-IDM achieves $92\%$ latent monotonicity (the agent's latent distance to the goal decreases at every step in $92\%$ of episodes) versus $73\%$ for CEM, and $35.5\times$ lower action jerk ($0.055$ vs.\ $1.962$).} 
\label{fig:tworoom_exec}
\end{figure}

Three \textcolor{black}{synergistic} mechanisms \textcolor{black}{drive} this result (Figure~\ref{fig:tworoom_exec} illustrates the effect on Two-Room). First, GC-IDM operates on real encoder outputs \textcolor{black}{during} both training and test time, eliminating the distribution shift that cripples \textcolor{black}{pairwise IDMs} (Appendix~\ref{app:pairwise}). Second, re-encoding the current observation at every step delivers \textcolor{black}{robust} closed-loop correction: each single-step error is overwritten by a fresh observation rather than \textcolor{black}{being} integrated over a commit window. Third, even \textcolor{black}{with a} low pointwise $R^2$\textcolor{black}{,} the IDM learns the average direction toward the goal, and per-step replanning accumulates these directionally correct predictions into successful trajectories. The ablations in Section~\ref{sec:ablations_main} test each mechanism independently; Appendix~\ref{app:why_gcidm} \textcolor{black}{provides an} analysis with error-propagation bounds. 

\subsection{Comparison with MPPI, iCEM, and Gradient-Based Planners}
\label{sec:solver_ablation}

\begin{figure}[t]
\centering
\includegraphics[width=\linewidth]{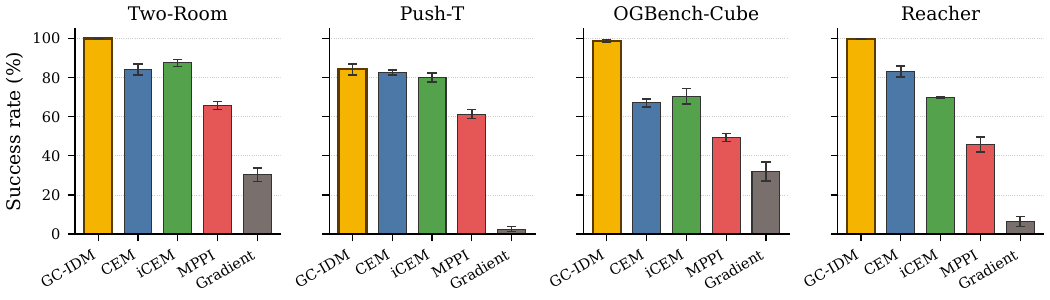}
\caption{\textbf{Solver-family comparison, $n{=}200$, four environments.} Success rate and per-plan-call wall-clock, mean $\pm$ std over three training seeds. All sampling solvers use \texttt{stable\_worldmodel} defaults; GradientSolver uses SGD with \texttt{lr}$=1.0$. GC-IDM is the highest-success method in every environment: it exceeds the best sampling baseline by $12.5$ pp on Two-Room, $1.7$ pp on Push-T, $28.2$ pp on Cube, and $29.4$ pp on Reacher, at $29$--$134\times$ lower per-plan-call cost. Appendix~\ref{app:solver_ablation_full} gives the $n{=}50$ protocol, per-episode timing, and full configuration details.}
\label{fig:solver_ablation}
\end{figure}

A compute sweep over 12 CEM configurations spanning a $500\times$ budget range shows that no configuration is simultaneously faster and more successful than GC-IDM on any environment (Figure~\ref{fig:pareto_grid}). To determine whether this gap extends beyond CEM's specific sampling scheme, we evaluated all four environments \textcolor{black}{using} three alternative solvers from \texttt{stable\_worldmodel} alongside CEM, each at the library's default hyperparameters: MPPI (softmax-weighted path-integral sampling)~\citep{williams2017mppi}, iCEM (CEM with temporally correlated noise and elite reuse)~\citep{pinneri2021icem}, and GradientSolver (SGD backpropagation through the world-model predictor). All four sampling-based planners share the same frozen LeWM encoder, the same \texttt{PlanConfig}, and identical episode samples per $(n, \text{seed})$ cell. Figure~\ref{fig:solver_ablation} reports the $n{=}200$ summary; the full per-protocol table with $n{=}50$ rows and wall-clock timing is \textcolor{black}{available} in Appendix~\ref{app:solver_ablation_full}.

At matched library defaults, GC-IDM \textcolor{black}{remains} the highest-success method in every $(n, \text{env})$ cell. The best sampling baseline differs by environment: iCEM on Two-Room ($87.5\%$) and Cube ($70.5\%$), \textcolor{black}{and} CEM on Push-T ($82.5\%$) and Reacher ($70.3\%$)\textcolor{black}{; however,} none of them closes the \textcolor{black}{performance} gap. The first-order GradientSolver collapses \textcolor{black}{across the board}, reaching only $2.5$--$32.0\%$ across the four environments; the default SGD step \textcolor{black}{fails to} recover trajectory-optimization-quality actions through the LeWM predictor \textcolor{black}{within} $30$ iterations. Speedups hold across the entire family: GC-IDM is $103$--$134\times$ faster per plan call than iCEM, $29$--$34\times$ faster than MPPI, and $2.0$--$2.4\times$ faster than Gradient\textcolor{black}{Solver}, \textcolor{black}{further complementing} the $104$--$130\times$ gap over CEM documented in Table~\ref{tab:headline}.

\subsection{Ablation Studies}
\label{sec:ablations_main}

\paragraph{Compute frontier.}
Sweeping CEM across $12$ configurations ($\texttt{num\_samples} \in \{30, 100, 300, 1000\}$, $\texttt{n\_steps} \in \{2, 5, 10, 30\}$; $60$--$30{,}000$ total rollouts per plan call), no \textcolor{black}{single} CEM configuration is simultaneously faster \emph{and} more successful than GC-IDM on any environment: the upper-left win region is empty across the full $500\times$ compute sweep. On Push-T, the best CEM configuration ($90\%$ at $11.4$\,s per episode) exceeds GC-IDM's $84.2\%$ but at $35\times$ the planning cost. On the other three environments, CEM saturates below GC-IDM regardless of \textcolor{black}{the allocated} budget: Two-Room at $86\%$, OGBench-Cube at $68\%$, \textcolor{black}{and} Reacher at $76\%$. The full Pareto plots are \textcolor{black}{provided} in Figure~\ref{fig:pareto_grid}.

\paragraph{Goal-distance robustness.}
Table~\ref{tab:goal_offset_main} \textcolor{black}{demonstrates} that GC-IDM \textcolor{black}{maintains strong performance even} as the start--goal offset increases. Two-Room \textcolor{black}{performance} is saturated at $100\%$ across the full range, \textcolor{black}{while} Reacher and OGBench-Cube remain nearly flat. Push-T degrades \textcolor{black}{more noticeably} with distance, consistent with the view that longer contact-rich action chains are the \textcolor{black}{primary regime where} local inverse recovery becomes difficult.

\paragraph{Horizon supervision.}
Table~\ref{tab:hmax_main} shows that \textcolor{black}{integrating} multi-step horizon supervision is necessary, especially on Push-T. Very short horizons \textcolor{black}{impair} performance, while gains largely saturate by $H_\text{max}{=}50$ at the $50$-step evaluation budget.

\begin{table*}[t]
\centering
\begin{minipage}[t]{0.54\linewidth}
\centering
\captionof{table}{{Ablation study on goal-distance.}}
\label{tab:goal_offset_main}
\small
\resizebox{\linewidth}{!}{%
\begin{tabular}{@{}lcccccc@{}}
\toprule
Goal offset (steps) & 5 & 10 & 15 & 25 & 35 & 50 \\
\midrule
Two-Room     & 100.0 & 100.0 & 100.0 & 100.0 & 100.0 & 100.0 \\
Reacher      & 99.5 & 100.0 & 99.5 & 99.5 & 99.0 & 99.0 \\
Push-T       & 94.0 & 90.5  & 88.5 & 84.5 & 76.0 & 70.0 \\
OGBench-Cube & 98.5 & 99.0  & 99.0 & 99.0 & 95.0 & 93.0 \\
\bottomrule
\end{tabular}
}
\end{minipage}\hfill
\begin{minipage}[t]{0.44\linewidth}
\centering
\captionof{table}{Ablation study on \textbf{$H_\text{max}$.}}
\label{tab:hmax_main}
\small
\resizebox{\linewidth}{!}{%
\begin{tabular}{@{}lccccc@{}}
\toprule
$H_\text{max}$ & 5 & 10 & 25 & \textbf{50} & 100 \\
\midrule
Two-Room & 81.5 & 99.0 & 100.0 & \textbf{100.0} & 100.0 \\
Push-T   & 46.5 & 67.5 & 84.0  & \textbf{85.0}  & 83.5 \\
Reacher  & 88.0 & 97.0 & 99.5  & \textbf{99.5}  & 99.5 \\
OGBench-Cube & 69.5 & 79.5 & 99.0 & \textbf{99.0} & 98.0 \\
\bottomrule
\end{tabular}
}
\end{minipage}
\end{table*}

\subsection{Trajectory Quality: Smoothness and Directness}
\label{sec:trajectory_quality}

Beyond \textcolor{black}{evaluating} binary success rate\textcolor{black}{s}, GC-IDM also produces qualitatively better trajectories than CEM. Two metrics capture complementary aspects of trajectory quality: \emph{action smoothness} (jerk) and \emph{directness of approach} in latent space (monotonicity).

\paragraph{Metrics.}
\emph{Action jerk} is the mean second finite difference of the action sequence: $\frac{1}{T-2}\sum_{t=2}^{T}\|\va_t - 2\va_{t-1} + \va_{t-2}\|_2$. Lower jerk \textcolor{black}{corresponds to} smoother, more physically plausible control. \textcolor{black}{Whereas} CEM replans in open-loop $5$-step blocks, creating discontinuities at block boundaries\textcolor{black}{,} GC-IDM acts closed-loop \textcolor{black}{at} every step with actions inherited from expert demonstrations via the IDM training set.
\emph{Latent monotonicity} measures $\|\vz_t - \vz_\text{goal}\|_2$ at every step and reports the fraction of episodes in which this distance decreases monotonically, meaning the agent moves steadily toward the goal in latent space without backtracking. See supplementary videos for rollout comparisons across environments.

\begin{table}[t]
\centering
\caption{\textbf{Trajectory quality: action jerk and latent monotonicity.} Both methods evaluated on the same $50$ episodes per environment (seed $42$, goal offset $25$, budget $50$) from matched initial states and goals. Jerk is the mean $\ell_2$ norm of the second finite difference of the action sequence (lower is smoother). Monotonicity is the fraction of episodes in which latent distance to the goal decreases at every step (higher is more direct).}\vspace{1mm}
\label{tab:trajectory_quality}
\small
\setlength{\tabcolsep}{5pt}
\begin{tabular}{@{}lcccccc@{}}
\toprule
 & \multicolumn{3}{c}{Action Jerk $\downarrow$} & \multicolumn{2}{c}{Monotonicity $\uparrow$} \\
\cmidrule(lr){2-4} \cmidrule(lr){5-6}
Environment & GC-IDM & CEM & Ratio & GC-IDM & CEM \\
\midrule
Two-Room & $\mathbf{0.055}$ & $1.962$ & $35.5\times$ & $\mathbf{92\%}$ & $73\%$ \\
Reacher  & $\mathbf{0.043}$ & $1.103$ & $25.7\times$ & $\mathbf{80\%}$ & $63\%$ \\
OGB-Cube & $\mathbf{0.083}$ & $1.721$ & $20.7\times$ & $69\%$ & $\mathbf{71\%}$ \\
Push-T   & $\mathbf{0.028}$ & $0.429$ & $15.3\times$ & $\mathbf{100\%}$ & $71\%$ \\
\bottomrule
\end{tabular}
\end{table}

Table~\ref{tab:trajectory_quality} reports both metrics on matched episodes. GC-IDM produces $15$--$36\times$ smoother actions across all four environments, and approaches the goal more directly in latent space on three of \textcolor{black}{the} four tasks (Cube \textcolor{black}{results in} a near-tie at $69\%$ vs.\ $71\%$). The smoothness advantage is structural: closed-loop single-step imitation inherits \textcolor{black}{the continuity of} expert-demonstration\textcolor{black}{s}, while CEM's open-loop replanning injects block-boundary discontinuities every $25$ raw environment steps. The monotonicity advantage is a \textcolor{black}{direct} consequence of re-encoding at every step: each action is chosen from the true current latent state, avoiding the drift that accumulates when CEM commits to an open-loop block from a predicted latent trajectory. 

\section{Conclusion and Future Work}
\label{sec:conclusion}

We propose GC-IDM, a learned planning method that eliminates \textcolor{black}{the need for} iterative test-time search in \textcolor{black}{pretrained} JEPA world model\textcolor{black}{s}. \textcolor{black}{While} LeWM is already an efficient world model, \textcolor{black}{featuring} ${\sim}15$M parameters, trainable on a single GPU in a few hours, \textcolor{black}{and} planning $48\times$ faster than foundation-model-based alternatives~\citep{lewm2026}, its CEM planner still dominates \textcolor{black}{the overall} inference cost. GC-IDM removes this bottleneck: a $1.5$M-parameter MLP trained in ${\sim}20$ minutes per environment on frozen LeWM embeddings replaces the entire $9{,}000$-rollout CEM loop with a single forward pass per step. \textcolor{black}{This approach} match\textcolor{black}{es} or exceed\textcolor{black}{s} CEM's success rate in seven of \textcolor{black}{the} eight environment--protocol cells while reducing planning cost by $100$--$130\times$. 

\textcolor{black}{Crucially, this performance advantage} extends to the broader family of test-time planners: MPPI, iCEM, and gradient-based optimization all fall below GC-IDM in \textcolor{black}{terms of} success rate \textcolor{black}{despite incurring} substantially higher computational cost\textcolor{black}{s}. These results support a simple and direct interpretation: when a world model's latent space is sufficiently well structured by regularization, the planning problem can be shifted \textcolor{black}{away} from online optimization \textcolor{black}{and completely} into learned inference. An important direction for future work is to test how far this regime extends to settings with stronger irreversibility, longer-horizon contact structure\textcolor{black}{s}, higher-dimensional action spaces, and different latent architectures. Additionally, we aim to explore the role of the regularizer itself by comparing isotropic and non-isotropic latent spaces on the same environments.


\bibliographystyle{plainnat}
\bibliography{references}


\clearpage
\appendix
\setcounter{table}{0}
\setcounter{figure}{0}
\setcounter{section}{0}
\renewcommand{\thetable}{\Alph{table}}
\renewcommand{\thefigure}{\Alph{figure}}
\renewcommand{\thesection}{\Alph{section}}
\label{app:qualitative}

\section{Why GC-IDM Works: Mechanism Analysis}
\label{app:why_gcidm}

 Here we develop each in detail and provide the error-propagation analysis that underpins the closed-loop argument.

\paragraph{Error propagation and replan frequency.}
Both CEM and GC-IDM are model predictive controllers with a receding horizon, and both are therefore closed-loop in the MPC sense. What differs is the interval between re-encodings. GC-IDM re-encodes the current observation and re-computes the action every environment step; LeWM's default CEM configuration (\texttt{receding\_horizon=5}, \texttt{action\_block=5}) commits $25$ raw environment actions before re-encoding, so a $50$-step episode contains $50$ re-encodings for GC-IDM and $2$ for CEM.

Let $\epsilon_t = \hat{\va}_t - \va_t^{*}$ denote the single-step action error at time $t$, where $\va_t^{*}$ is the optimal action toward the goal, and let $\delta_{t+1}$ denote the resulting observation divergence. Inside any window of $T$ environment steps during which the controller commits actions without re-encoding, divergence accumulates:
\begin{equation}
   \delta_T^{\text{open}} \;\le\; L \sum_{t=1}^{T} \|\epsilon_t\|,
   \label{eq:open_loop_error}
\end{equation}
where $L$ is the Lipschitz constant of the environment transition with respect to actions. A controller that re-encodes after every step and recomputes $\hat{\va}_{t+1}$ from the fresh observation satisfies the tighter per-step bound
\begin{equation}
   \delta^{\text{closed}}_{t+1} \;\le\; L \cdot \|\epsilon_t\|,
   \label{eq:closed_loop_error}
\end{equation}
because each step's error is overwritten by the next re-encoding rather than integrated forward over additional committed actions. GC-IDM operates under~(\ref{eq:closed_loop_error}) at every step; CEM operates under~(\ref{eq:open_loop_error}) with $T{=}25$ inside each commit window and under~(\ref{eq:closed_loop_error}) across window boundaries.

\paragraph{Three mechanisms in detail.}
GC-IDM achieves $100\%$ success on Two-Room despite a training $R^2$ of only $0.20$. Three mechanisms account for this:

\emph{Mechanism 1: No train/test distribution shift.} GC-IDM operates on real ViT-encoded observations at both training and test time, eliminating the distribution shift that forces the pairwise IDM (Appendix~\ref{app:pairwise}) into extreme noise augmentation ($\sigma{=}1.5$). This is confirmed by the noise ablation (Table~\ref{tab:abl_noise_gcidm}): every environment prefers $\sigma{=}0$.

\emph{Mechanism 2: Closed-loop correction via re-encoding.} Re-encoding the current observation at every environment step delivers the error bound in Equation~(\ref{eq:closed_loop_error}): any single-step prediction error is overwritten one step later by a fresh observation. This mechanism predicts that GC-IDM should be insensitive to noise injection during training, because at test time it sees clean encoder outputs anyway, exactly what Table~\ref{tab:abl_noise_gcidm} shows.

\emph{Mechanism 3: Directionally correct predictions accumulate.} Even at low $R^2$ the IDM learns the average direction toward the goal at each state, and per-step replanning accumulates these noisy but directionally correct predictions into successful trajectories, similar in spirit to potential-field navigation~\citep{khatib1986potentialfield}. The latent monotonicity measurements in Table~\ref{tab:trajectory_quality} provide direct evidence: on Two-Room, $92\%$ of GC-IDM episodes show monotonically decreasing latent distance to the goal at every step. The same effect appears in the bias--variance decomposition of \citet{pidm_theory2026}: future-state conditioning trades a small predictor bias for a large variance reduction in the action estimator, and a low-$R^2$ pointwise estimator can therefore induce a high-quality closed-loop policy.

Together these three mechanisms replace search over action sequences with a directed step in one forward pass: at each environment step, GC-IDM commits to the action that moves the latent state toward the goal under its current estimate, the environment evolves, and a fresh encoding replaces the estimate for the next step. No candidate trajectories are ever enumerated or scored. Each mechanism implies a testable empirical prediction: closed-loop re-encoding $\to$ noise insensitivity (Table~\ref{tab:abl_noise_gcidm}); horizon supervision $\to$ removal collapses performance even at higher capacity (Table~\ref{tab:abl_horizon_embed}); smooth latent metric $\to$ no benefit from extra capacity (Table~\ref{tab:abl_arch}).

\section{Theoretical Analysis: Isotropy and Inverse-Dynamics Conditioning}
\label{app:theory}

Under idealized conditions on the latent geometry, the inverse-dynamics problem has a favorable first-order form. This analysis is \emph{conditional}: it identifies geometric properties that would make inverse dynamics easier to learn, but we have not verified that LeWM's latent space satisfies these conditions in full, nor have we ablated against a non-isotropic baseline. The empirical results in the main paper stand independently of this theoretical framing.

Let $s_t$ denote the underlying state, $a_t$ the action, and $z_t = e(s_t)$ the latent embedding produced by the frozen encoder. Consider the one-step latent transition map
\begin{equation}
    F(z_t, a_t) \;\triangleq\; e\bigl(f(s_t, a_t)\bigr),
    \label{eq:latent_transition_map}
\end{equation}
where $f$ is the environment dynamics and $s_t = e^{-1}(z_t)$ is understood only locally on the data manifold.

\begin{assumption}[Isotropic latent statistics]
On the demonstration distribution the latent variable is approximately whitened and isotropic in the second-order sense,
\begin{equation}
    \E[z_t] = 0,
    \qquad
    \mathrm{Cov}(z_t) = I,
    \label{eq:isotropic_gaussian_assumption}
\end{equation}
with the additional modeling idealization that the local latent density is close to Gaussian on the region visited by demonstrations.
\end{assumption}

\begin{assumption}[Local regularity and action identifiability]
Near the demonstration manifold: (i) $e$ is differentiable and locally bi-Lipschitz; (ii) $f$ is differentiable in $(s,a)$; and (iii) for fixed $z_t$, the Jacobian $\partial F(z_t,a)/\partial a$ has full column rank.
\end{assumption}

Under these assumptions, linearizing at $(s_t,a_t)$ gives, for a nearby action perturbation $\delta a$,
\begin{equation}
    F(z_t, a_t + \delta a)
    \;\approx\;
    F(z_t, a_t) + B_t \, \delta a,
    \qquad
    B_t \;\triangleq\; \frac{\partial F(z_t,a_t)}{\partial a}.
    \label{eq:linearized_latent_dynamics}
\end{equation}
Defining the induced latent displacement $\delta z \triangleq F(z_t,a_t+\delta a)-F(z_t,a_t)$, the local inverse problem becomes $\delta z \approx B_t \, \delta a$. When $B_t$ has full column rank, the minimum-norm local solution is
\begin{equation}
    \delta a \approx B_t^{\dagger} \, \delta z
    \;=
    (B_t^\top B_t)^{-1} B_t^\top \delta z,
    \label{eq:pseudoinverse_inverse}
\end{equation}
where $B_t^{\dagger}$ is the Moore--Penrose pseudoinverse.

\begin{proposition}[Local inverse map]
Under Assumption~2, the implicit function theorem yields a local inverse map $G$ satisfying
\begin{equation}
    a_t = G(z_t, z_{t+1}),
    \qquad
    z_{t+1} = F\bigl(z_t, G(z_t, z_{t+1})\bigr),
    \label{eq:local_inverse}
\end{equation}
and this inverse is continuous (indeed differentiable) in a neighborhood of each training point.
\end{proposition}

In words, as long as actions induce locally distinguishable motion in latent space, inverse dynamics is a well-posed local regression problem.

\paragraph{Why isotropy would improve conditioning.}
The learnability question is controlled by the singular spectrum of $B_t = \partial F/\partial a$. By the chain rule,
\begin{equation}
    B_t
    \;=
    J_e(s_{t+1}) \, J_f^{(a)}(s_t,a_t),
    \label{eq:chain_rule_B}
\end{equation}
where $J_e$ is the encoder Jacobian and $J_f^{(a)}$ is the dynamics Jacobian with respect to action. If, in addition, the encoder is locally near-isometric in the sense that
\begin{equation}
    m I \;\preceq\; J_e(s)^\top J_e(s) \;\preceq\; M I
    \qquad \text{with } M/m \text{ small,}
    \label{eq:near_isometry}
\end{equation}
then the singular values of $J_e(s)$ all lie in $[\sqrt{m}, \sqrt{M}]$, so
\begin{equation}
    \kappa(B_t)
    \;\le\;
    \kappa\!\bigl(J_e(s_{t+1})\bigr)
    \, \kappa\!\bigl(J_f^{(a)}(s_t,a_t)\bigr)
    \;\le\;
    \sqrt{M/m} \, \kappa\!\bigl(J_f^{(a)}(s_t,a_t)\bigr).
    \label{eq:conditioning_bound}
\end{equation}
Thus, when the encoder is close to isotropic at the Jacobian level, it does not introduce substantial additional ill-conditioning beyond that already present in the action-to-state map. SIGReg fits this picture by discouraging collapsed or disproportionately amplified latent directions. The rigorous claim is conditional: distributional isotropy alone does not imply Equation~(\ref{eq:near_isometry}), but if LeJEPA-style isotropic statistics are accompanied by local near-isometry of the encoder, then the induced inverse problem satisfies Equation~(\ref{eq:conditioning_bound}) and is easier to approximate with a small model.

\section{More Qualitative Comparisons}

\begin{figure}[ht]
\centering
\includegraphics[width=\linewidth]{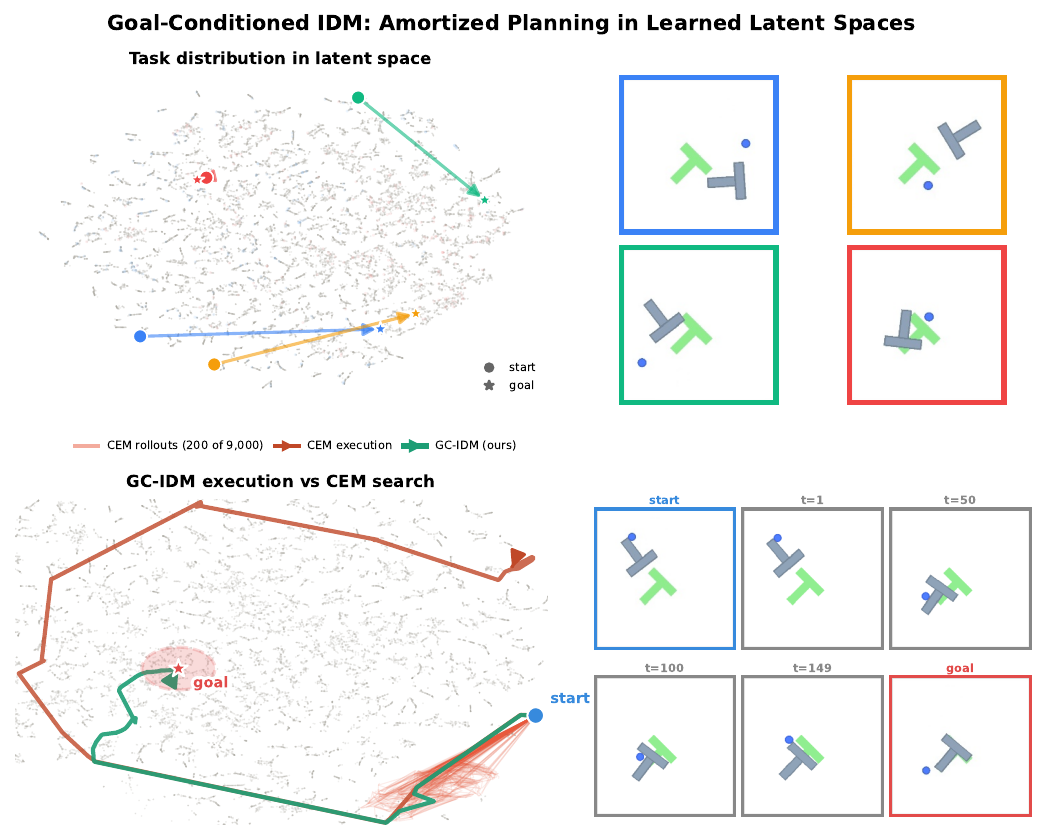}
\caption{\textbf{Comparisons on latent trajectory rollouts.} \textbf{\textit{Top}}: Four goal-conditioned tasks in the PushT environment, visualized in the 192-dimensional LeWM latent space (t-SNE projection; grey points are 20K training embeddings). Each colored circle marks a different starting configuration (right panels: the grey T-block and blue agent vary across tasks), while stars mark the corresponding goal observations. All four tasks share the same physical target (the green T-block at a fixed position), but goal embeddings differ because the encoder represents the full scene, including agent and block positions at the goal timestep, not the target alone. \textbf{\textit{Bottom}}: Execution comparison on a single task. Light red lines visualize 200 of the 9,000 random candidate rollouts evaluated per CEM plan call, each spanning a 5-step prediction horizon; CEM replans every 25 environment steps. The dark red path shows CEM's actual executed trajectory; the green path shows GC-IDM (ours). GC-IDM produces one action per step via a single forward pass through the inverse dynamics model, requiring no iterative search. Right panels show the GC-IDM execution progression from start to goal.}
\label{fig:qualitative_rollouts}
\end{figure}

\paragraph{Comparison on latent planning trajectories.} Figure~\ref{fig:qualitative_rollouts} provides a qualitative view of GC-IDM and CEM rollouts across representative episodes. On Two-Room and Reacher, both methods produce visually coherent trajectories, but GC-IDM's per-step action sequences are smoother because there is no re-optimization jitter from the CEM sampling loop. On Push-T and OGBench-Cube, the contact-rich phases reveal the main behavioral difference: GC-IDM commits to a single directed step per environment step and relies on re-encoding to correct, while CEM's action sequences show larger step-to-step variance from the repeated sampling-and-refinement cycle. When GC-IDM fails on Push-T, it typically undershoots the rotation component of the goal pose rather than diverging entirely, consistent with the view that longer contact sequences are the regime where local inverse recovery becomes harder.

\label{app:pareto}

\begin{figure}[h]
\centering
\includegraphics[width=\linewidth]{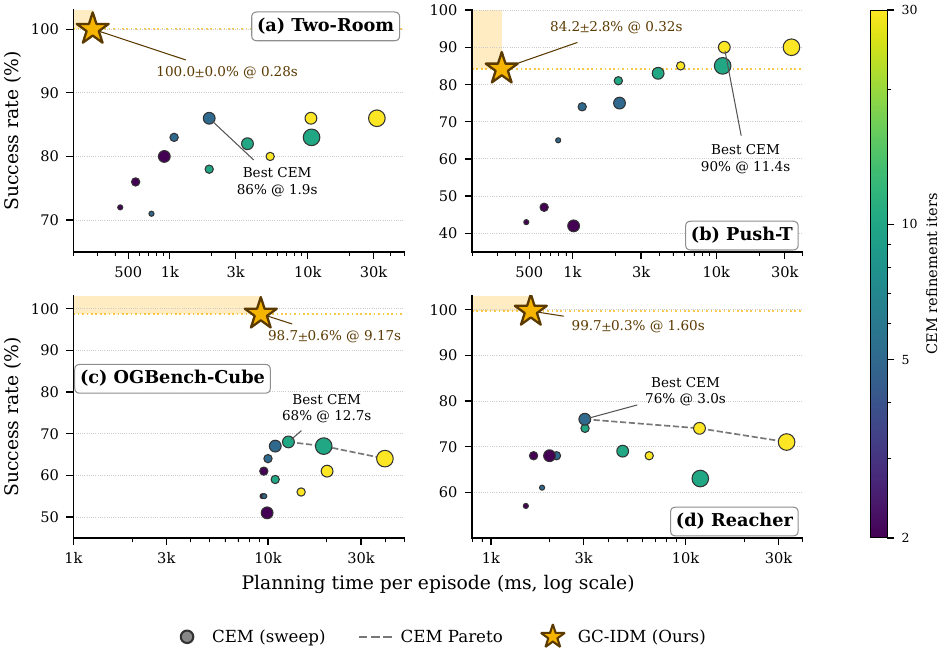}
\caption{\textbf{CEM compute vs.\ success rate across four LeWM environments.} }
\label{fig:pareto_grid}
\end{figure}

\paragraph{CEM pareto frontier analysis.} As is shown in Fig.\ref{fig:pareto_grid}, each panel sweeps CEM across $12$ configurations ($\texttt{num\_samples} \in \{30, 100, 300, 1000\}$, $\texttt{n\_steps} \in \{2, 5, 10, 30\}$; $60$--$30{,}000$ total rollouts), evaluated on $100$ in-distribution episodes per configuration at seed $42$. Dot color encodes refinement iterations and dot size encodes \texttt{num\_samples}. The dashed grey curve is the CEM upper-left Pareto envelope. The shaded yellow region marks configurations strictly faster and more successful than GC-IDM (gold star); it is empty in every panel across the full $500\times$ compute sweep. 


\begin{figure}[h]
\centering
\includegraphics[width=\linewidth]{figures/tworoom_exec_A.pdf}
\vspace{0.3em}
\includegraphics[width=\linewidth]{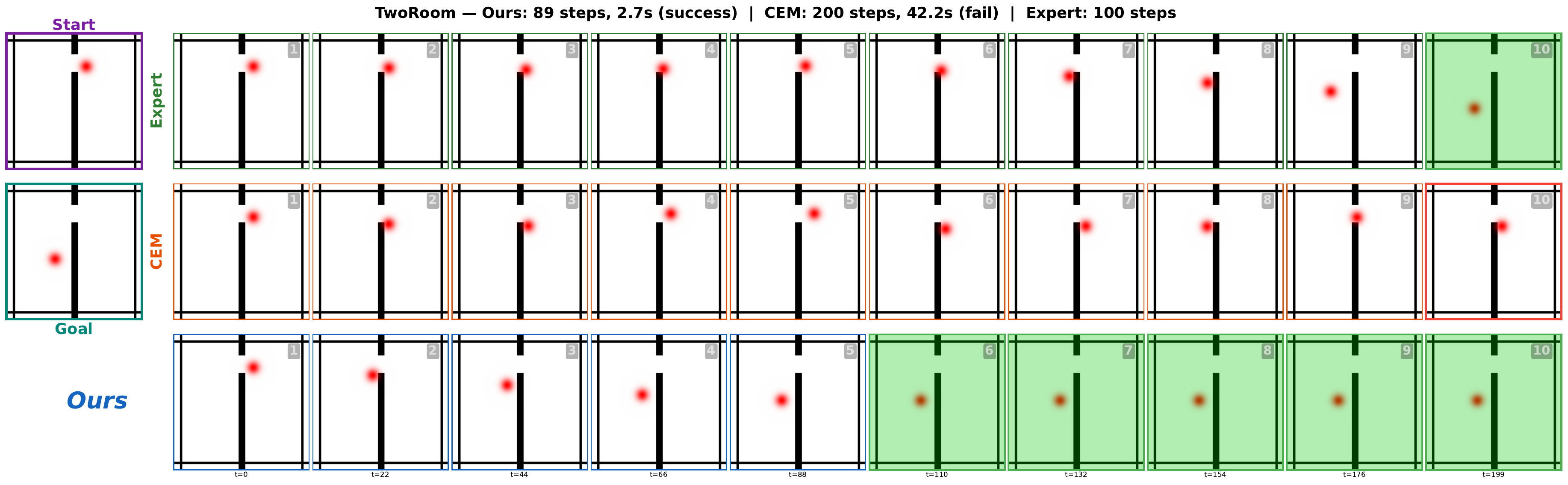}
\vspace{0.3em}
\includegraphics[width=\linewidth]{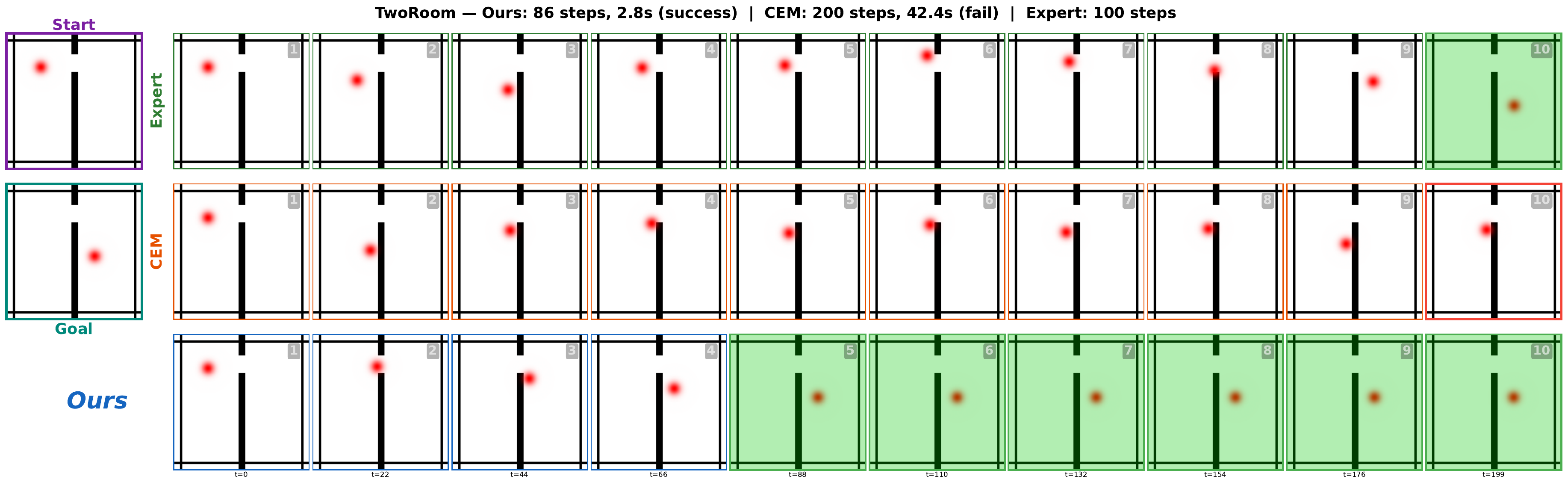}
\vspace{0.3em}
\includegraphics[width=\linewidth]{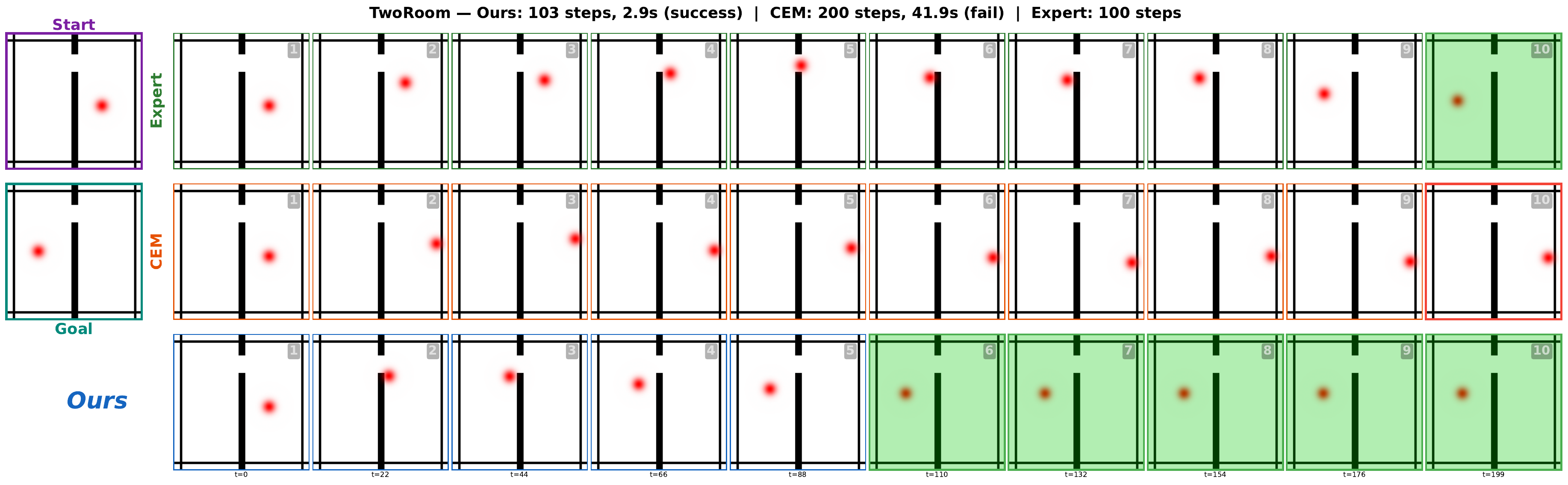}
\caption{\textbf{Two-Room: matched execution rollouts (4 episodes).}}
\label{fig:exec_tworoom}
\end{figure}

\begin{figure}[h]
\centering
\includegraphics[width=\linewidth]{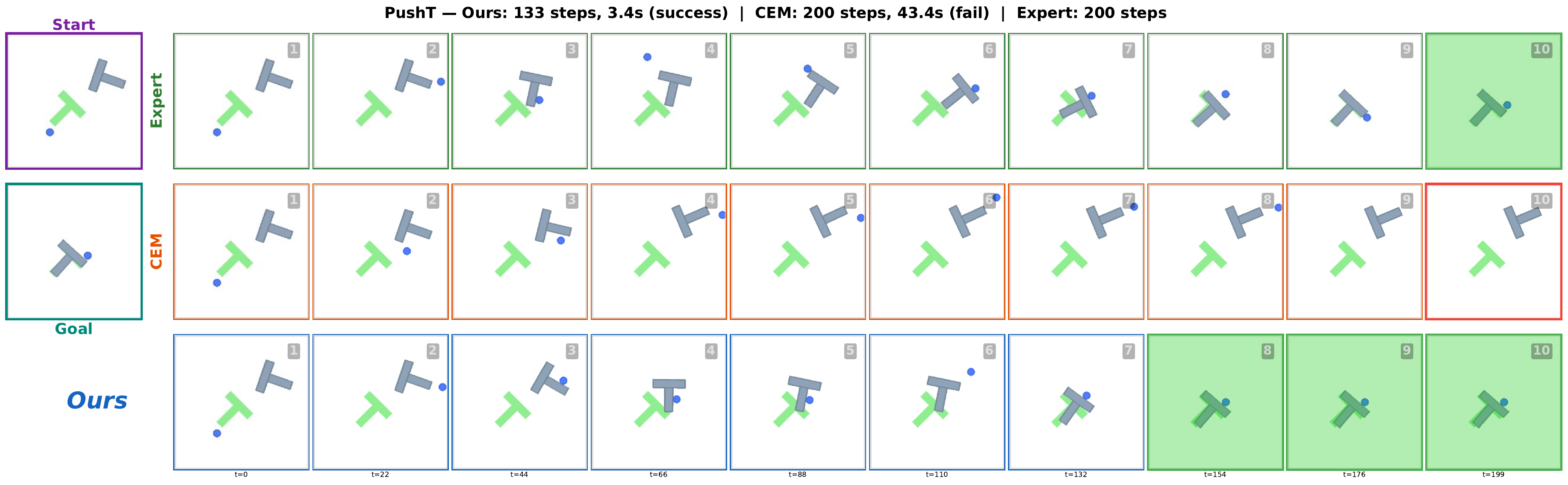}
\vspace{0.5em}
\includegraphics[width=\linewidth]{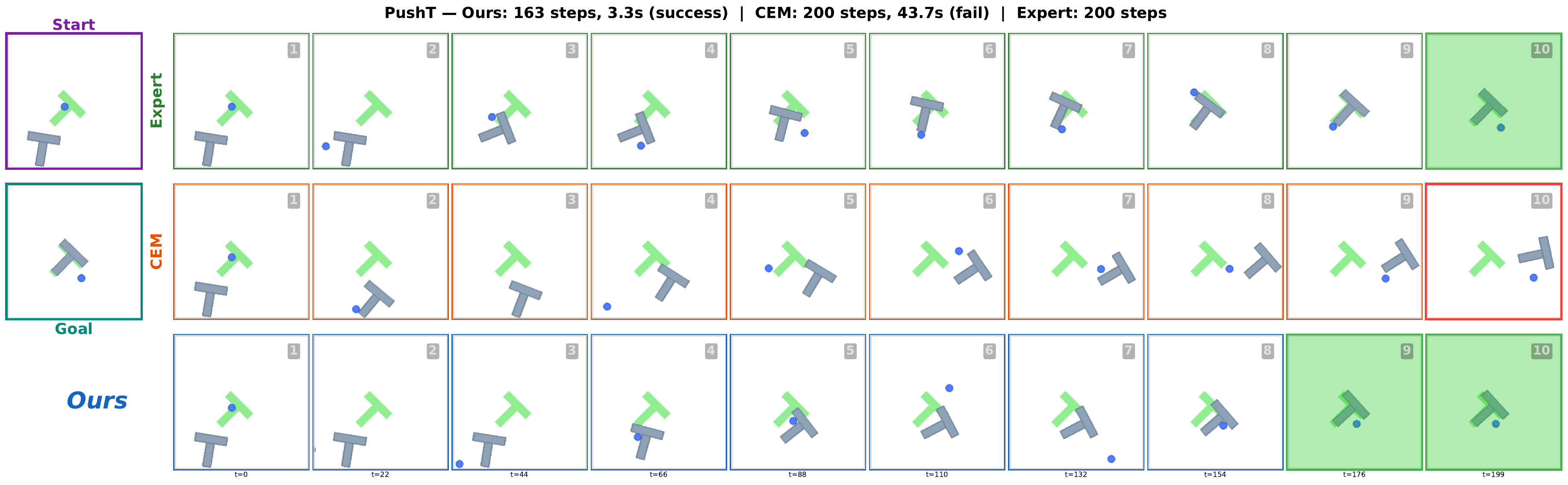}
\vspace{0.5em}
\includegraphics[width=\linewidth]{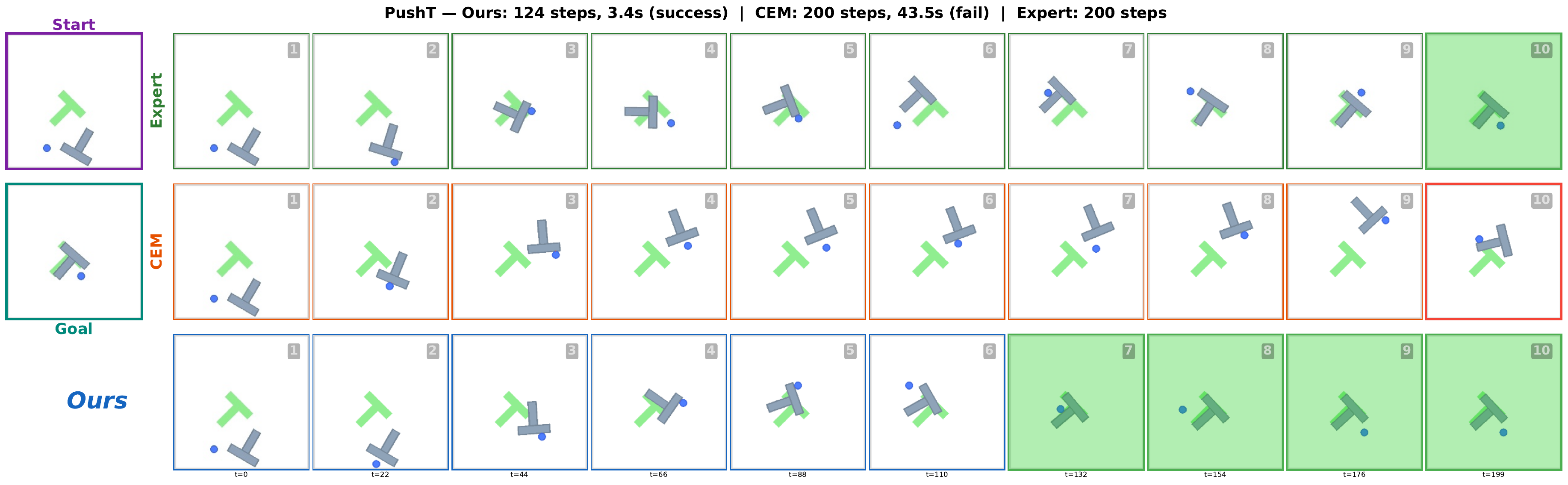}
\vspace{0.5em}
\includegraphics[width=\linewidth]{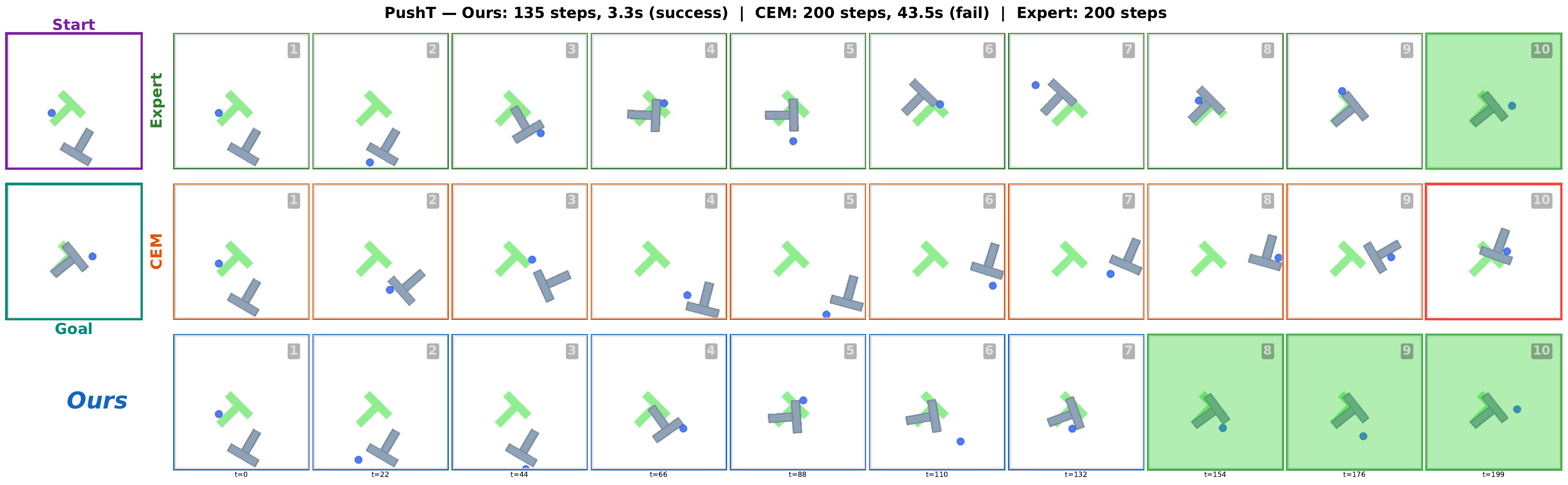}
\caption{\textbf{Push-T: matched execution rollouts (4 episodes).}}
\label{fig:exec_pusht}
\end{figure}

\begin{figure}[h]
\centering
\includegraphics[width=\linewidth]{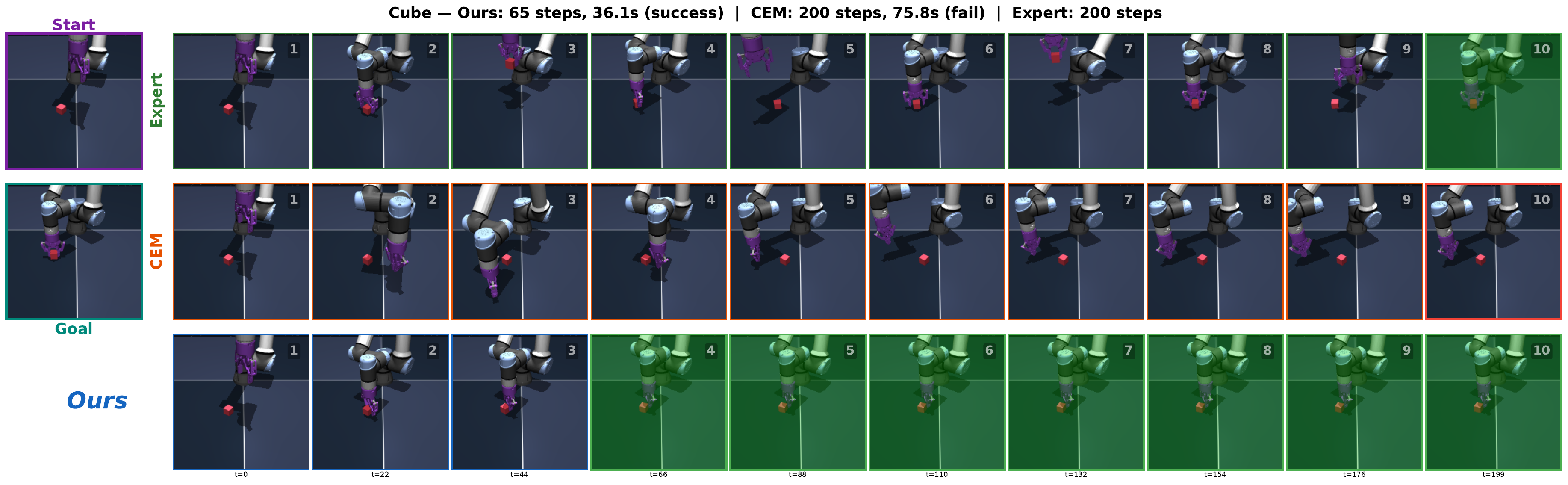}
\vspace{0.5em}
\includegraphics[width=\linewidth]{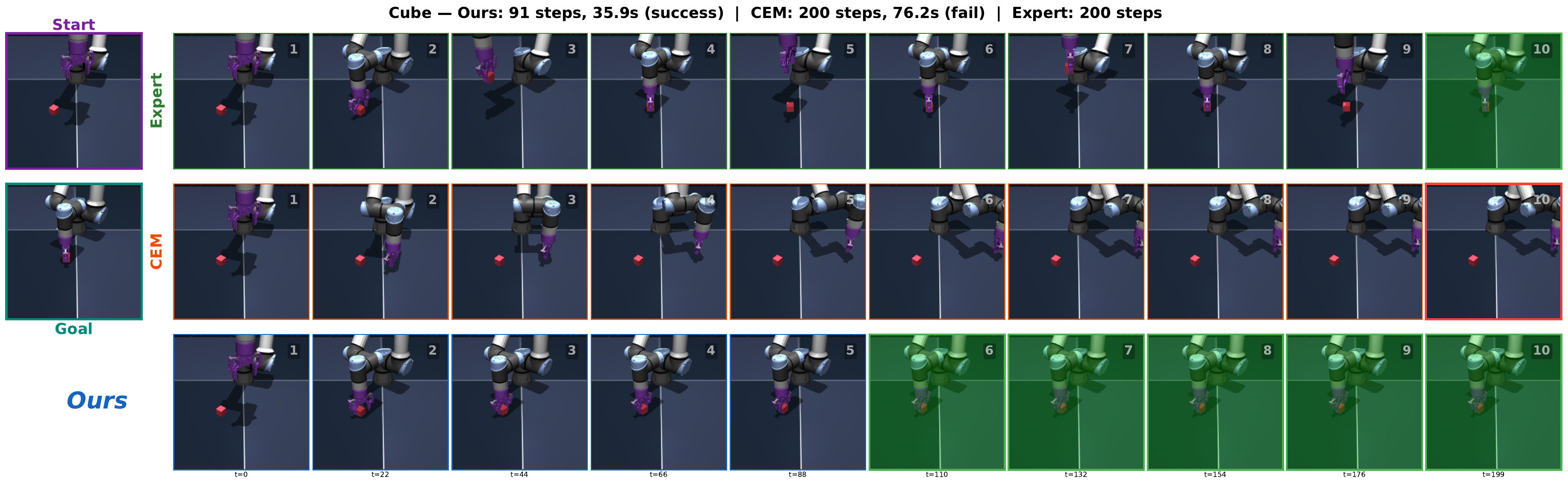}
\vspace{0.5em}
\includegraphics[width=\linewidth]{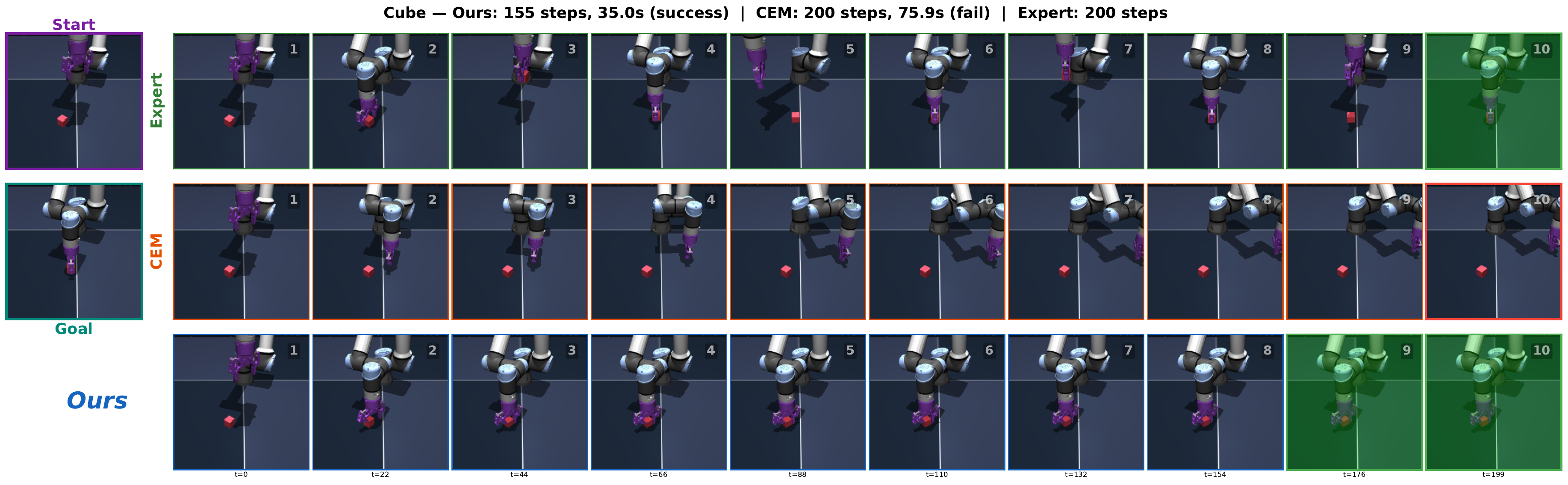}
\vspace{0.5em}
\includegraphics[width=\linewidth]{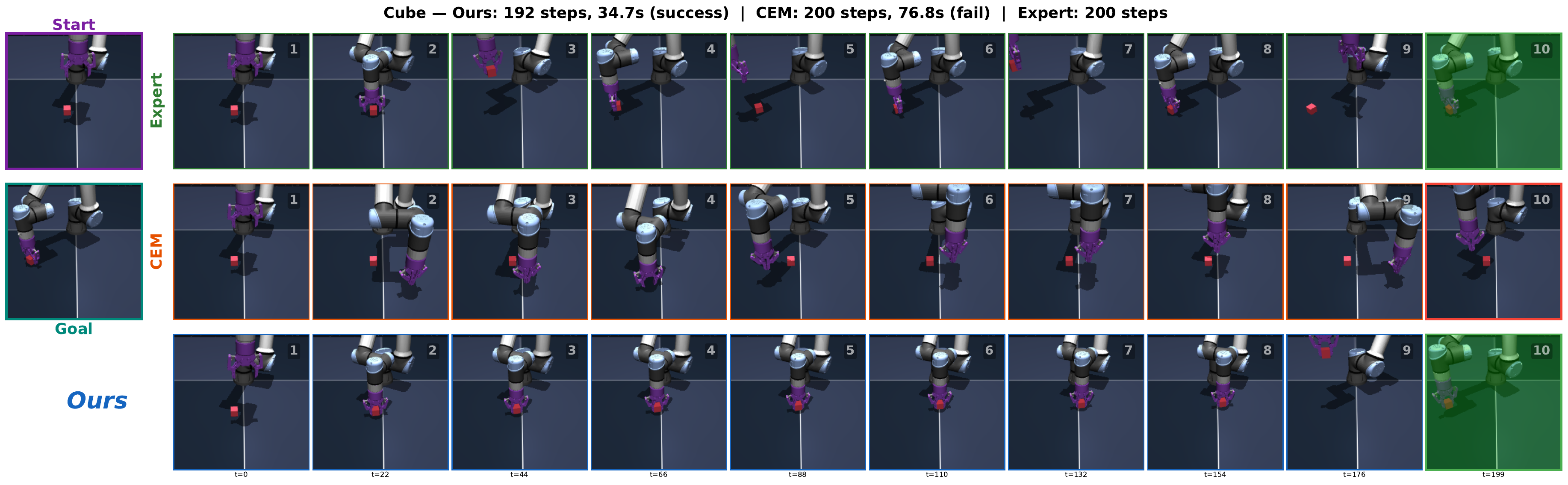}
\caption{\textbf{OGBench-Cube: matched execution rollouts (4 episodes).}}
\label{fig:exec_cube}
\end{figure}

\begin{figure}[h]
\centering
\includegraphics[width=\linewidth]{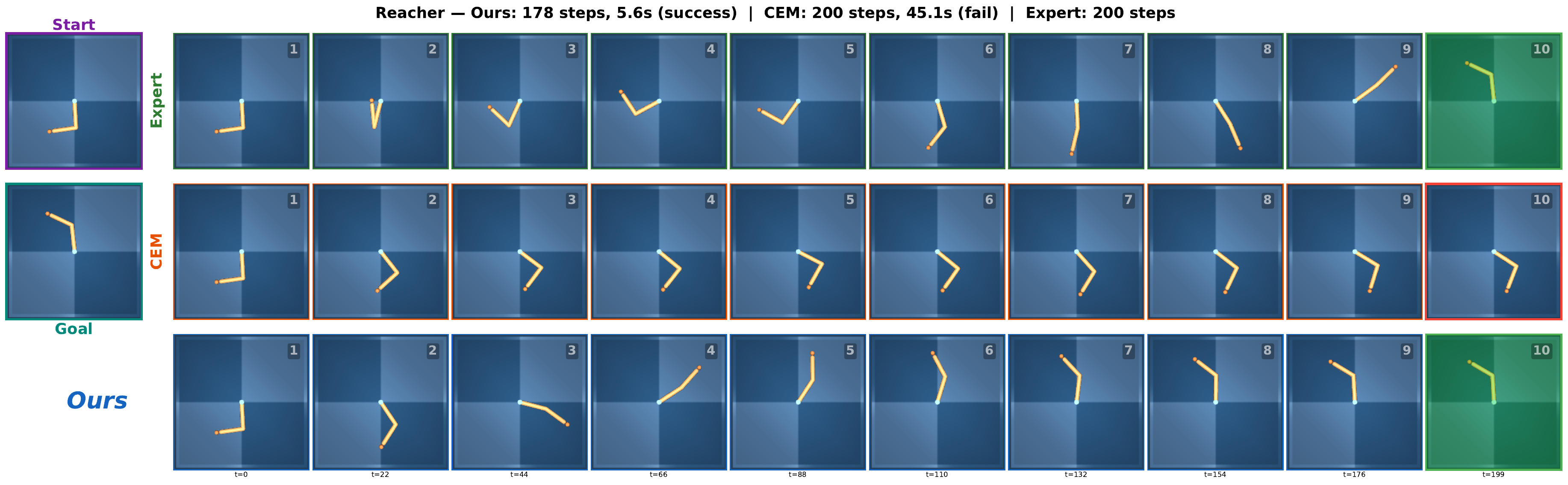}
\vspace{0.5em}
\includegraphics[width=\linewidth]{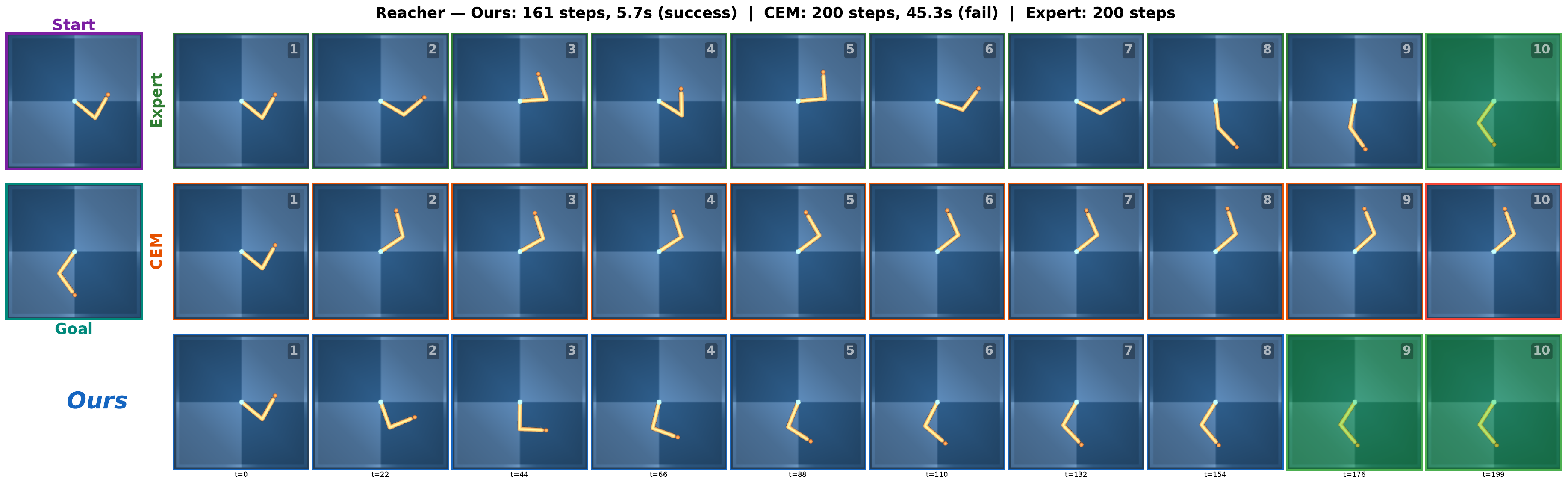}
\vspace{0.5em}
\includegraphics[width=\linewidth]{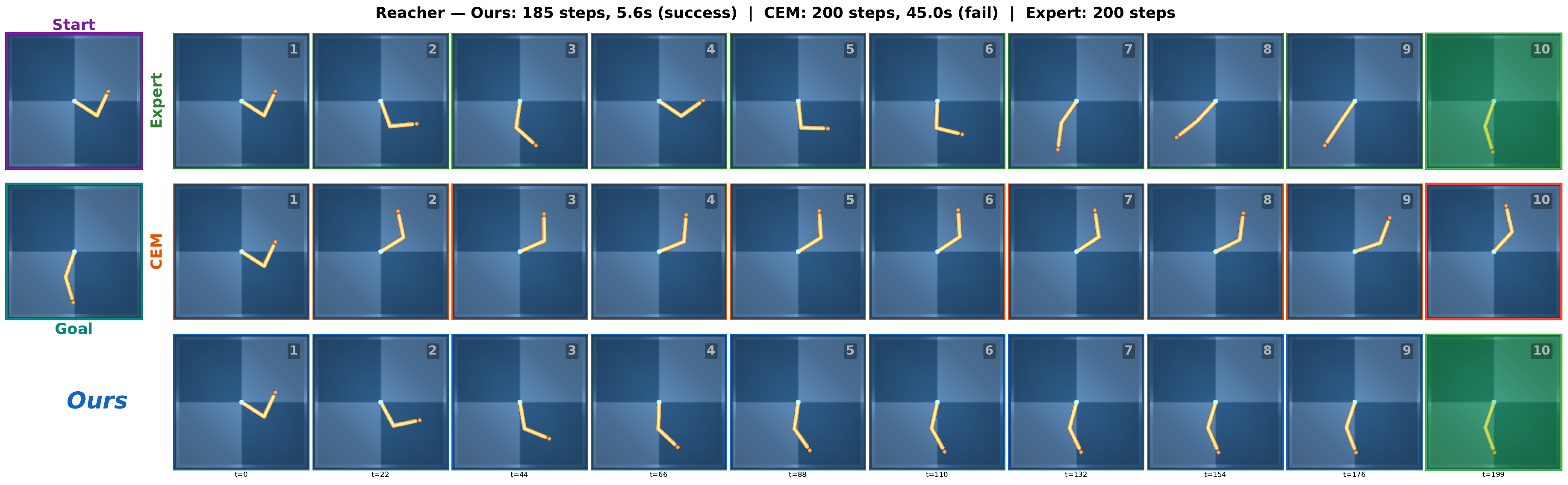}
\vspace{0.5em}
\includegraphics[width=\linewidth]{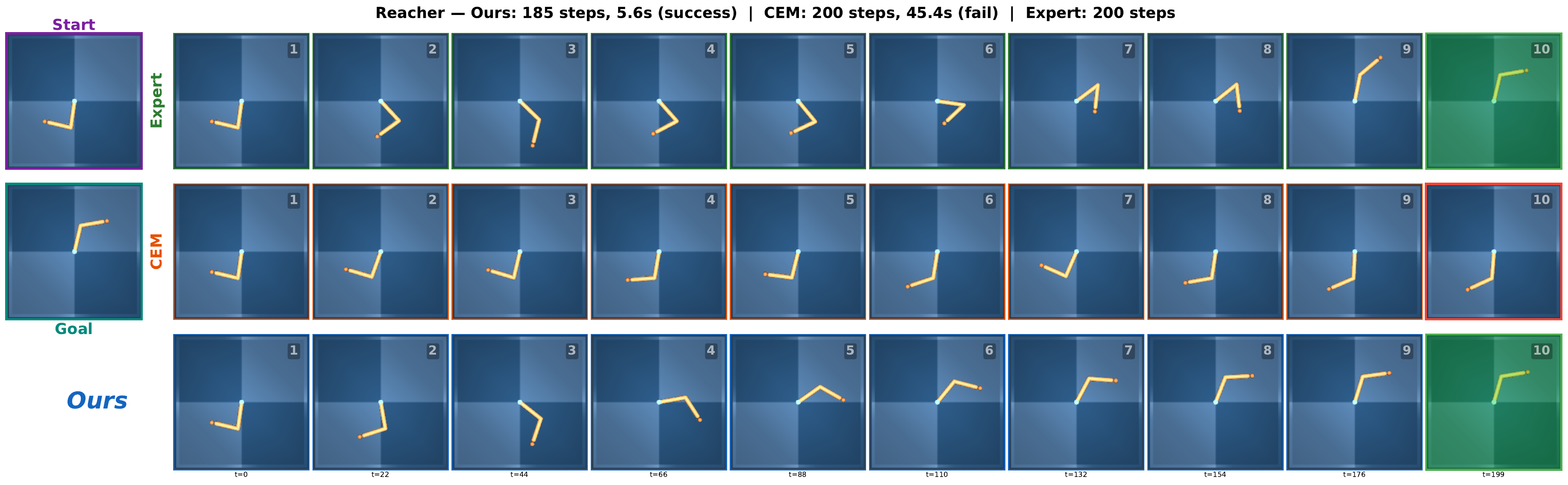}
\caption{\textbf{Reacher: matched execution rollouts (4 episodes).}}
\label{fig:exec_reacher}
\end{figure}

\label{app:exec_rollouts}
\paragraph{Comparisons on execution rollout frames.} Each figure below shows four matched episodes per environment: expert, CEM, and GC-IDM (ours) from identical start and goal states. The first column shows the start (top) and goal (bottom); columns 1--10 are evenly spaced frames. Green shading marks success; red borders mark failure. CEM and GC-IDM share the same time axis; the expert row uses the dataset's own time axis. Evaluation budget is $200$ steps; goal offset is $100$ for Two-Room and $200$ for the remaining three. Results are organized by environment: Two-Room (Figure~\ref{fig:exec_tworoom}), Push-T (Figure~\ref{fig:exec_pusht}), OGBench-Cube (Figure~\ref{fig:exec_cube}), and Reacher (Figure~\ref{fig:exec_reacher}).

\clearpage
\section{Pilot Study on Pairwise IDM}
\label{app:pairwise}

The pairwise IDM motivates the goal-conditioned formulation. Here we give the full pairwise method, the oracle experiment that was the original empirical anchor, the noise sweep, and the closed-loop failure analysis.

\subsection{Pairwise IDM}

The pairwise IDM learns the mapping $(\vz_t, \vz_{t+1}) \to \va_t$ from consecutive latent state pairs. Given a trained LeWM encoder, we extract embeddings for all frames in the training dataset and construct supervision triples $(\vz_t, \vz_{t+1}, \va_t)$ where $\va_t$ is the ground-truth action that produced the transition.

\paragraph{Architecture.}
The pairwise IDM is a 3-layer MLP with LayerNorm and GELU activations:
\begin{equation}
    \hat{\va}_t = \idm_\psi(\vz_t \,\|\, \vz_{t+1}),
    \label{eq:pairwise_idm}
\end{equation}
where $\|$ denotes concatenation. With $d{=}192$ (LeWM embedding dimension), hidden dimension $512$, and action dimension $2$, the model has ${\sim}730$K parameters.

\paragraph{Frameskip.}
LeWM trains with frameskip $k{=}5$, meaning the predictor operates on every $5$th frame. A natural choice is to train the IDM at the same temporal resolution, predicting concatenated action chunks $[\va_t, \ldots, \va_{t+k-1}]$ from pairs $(\vz_t, \vz_{t+k})$. However, this mapping is fundamentally ambiguous: many distinct $k$-step action sequences can connect the same two latent states. We find that frameskip-$5$ IDMs achieve $R^2 \approx 0$ on Two-Room. Training with frameskip-$1$ (single raw actions from consecutive frames) yields $R^2 = 0.993$, and we adopt this throughout.

\subsection{Noisy Conditioning}

A key insight from mimic-video~\citep{pai2025mimicvideo} is that the IDM must be robust to imperfect conditioning signals at test time. During training, the IDM receives ground-truth encoder embeddings; at test time it receives interpolated or predictor-generated embeddings, which differ in distribution. To bridge this gap, we add Gaussian noise to the input embeddings during training:
\begin{equation}
    \vz_\text{noisy} = \vz + \sigma \cdot \bm{\epsilon}, \quad \bm{\epsilon} \sim \mathcal{N}(\bm{0}, \bm{I}).
    \label{eq:noise}
\end{equation}

\subsection{Oracle Experiment and Noise Ablation}
\label{app:oracle}

The pairwise IDM receives ground-truth encoder embeddings during training but interpolated or predictor-generated embeddings at test time. Following mimic-video~\citep{pai2025mimicvideo}, we add Gaussian noise $\vz_\text{noisy} = \vz + \sigma \cdot \bm{\epsilon}$, $\bm{\epsilon} \sim \mathcal{N}(\bm{0}, \bm{I})$ to bridge this distribution gap. Two experiments characterize its effect on Two-Room. Table~\ref{tab:oracle} tests a single IDM trained at $\sigma{=}0$ against increasing test-time noise on ground-truth embedding pairs, with no environment interaction. At $\sigma{=}0$ reconstruction is near-perfect ($R^2{=}0.993$), confirming that LeWM's latent transitions encode essentially all the information needed to determine the action; reconstruction degrades gracefully up to $\sigma{\approx}0.5$ and collapses beyond $\sigma{=}1$. Table~\ref{tab:noise_ablation} trains eight separate IDMs at different $\sigma$ values and evaluates each closed-loop with lerp planning. The optimal training noise is $\sigma{=}1.5$ ($34\%$ success), corresponding to $150\%$ of the per-dimension embedding standard deviation, reflecting the large distribution shift between ground-truth and linearly interpolated embeddings. The gap between oracle quality ($R^2 = 0.993$) and the best closed-loop success ($34\%$) confirms that the bottleneck lies in trajectory generation, not action decoding.

\begin{table}[ht]
\centering
\begin{minipage}[t]{0.52\linewidth}
\centering
\captionof{table}{Oracle noise robustness: single $\sigma{=}0$ model, test-time noise on ground-truth pairs.}
\label{tab:oracle}
\small
\begin{tabular}{@{}lccc@{}}
\toprule
Test $\sigma$ & MSE & Cos Sim & $R^2$ \\
\midrule
0.00 & 0.005017 & 0.9966 & 0.9933 \\
0.01 & 0.005032 & 0.9966 & 0.9933 \\
0.02 & 0.005074 & 0.9966 & 0.9933 \\
0.05 & 0.005733 & 0.9961 & 0.9924 \\
0.10 & 0.008802 & 0.9936 & 0.9883 \\
0.20 & 0.018036 & 0.9860 & 0.9761 \\
0.50 & 0.081068 & 0.9416 & 0.8924 \\
1.00 & 0.321688 & 0.7303 & 0.5731 \\
2.00 & 0.720835 & 0.3286 & 0.0435 \\
\bottomrule
\end{tabular}
\end{minipage}\hfill
\begin{minipage}[t]{0.44\linewidth}
\centering
\captionof{table}{Training noise ablation: 8 separate models, closed-loop with lerp.}
\label{tab:noise_ablation}
\small
\begin{tabular}{@{}lcc@{}}
\toprule
Train $\sigma$ & Val MSE & Success \\
\midrule
0.0 & 0.005540 & 18\% \\
0.1 & 0.005986 & 20\% \\
0.2 & 0.006591 & 26\% \\
0.5 & 0.009219 & 24\% \\
0.8 & 0.011906 & 28\% \\
1.0 & 0.013480 & 26\% \\
\textbf{1.5} & \textbf{0.019367} & \textbf{34\%} \\
2.0 & 0.028369 & 28\% \\
\bottomrule
\end{tabular}
\end{minipage}
\end{table}

\subsection{Closed-Loop Performance and Failure Mode}
\label{app:pairwise_results}

Given an encoded start $\vz_\text{start}$ and goal $\vz_\text{goal}$, we generate a latent trajectory via linear interpolation $\vz_i = (1{-}\alpha_i)\vz_\text{start} + \alpha_i \vz_\text{goal}$ for $\alpha_i = i/H$, then decode actions from consecutive pairs using the pairwise IDM. Optionally, a refinement step rolls out the decoded actions through the predictor to obtain a corrected trajectory, which is then re-decoded; this can be iterated $K$ times. Table~\ref{tab:pairwise} reports the resulting performance on Two-Room.

\begin{table}[ht]
\centering
\caption{Pairwise IDM planning results. Linear interpolation (lerp) is the fundamental bottleneck: even with predictor-based refinement and multiple candidates, success rates plateau well below CEM. }
\label{tab:pairwise}
\begin{tabular}{@{}llccc@{}}
\toprule
Environment & Configuration & Success & ms/plan & Speedup \\
\midrule
\multirow{5}{*}{Two-Room}
    & Lerp only ($\sigma{=}1.5$)        & 34\% & 64   & $16\times$ \\
    & + Refine$\times$1                  & 36\% & 119  & $8\times$ \\
    & + Refine$\times$3                  & 42\% & 285  & $3.5\times$ \\
    & + Refine$\times$3, 10 candidates   & 48\% & 195  & $5\times$ \\
    & CEM (baseline)                     & 87\% & 1000 & $1\times$ \\
\bottomrule
\end{tabular}
\end{table}

\paragraph{Linear interpolation fails on complex tasks.}
The pairwise IDM with linear interpolation achieves at most $48\%$ on Two-Room (vs.\ CEM's $87\%$). The failure mode is clear: linear interpolation assumes the optimal latent path between start and goal is a straight line. For Two-Room, this assumption fails because the path must curve through the doorway. 

\paragraph{Refinement helps but saturates.}
Predictor-based refinement (rolling out decoded actions through the predictor to correct the trajectory) improves success from $34\%$ to $42\%$ on Two-Room. Additional iterations beyond $3$ yield diminishing returns as predictor errors compound. Adding $10$ noisy trajectory candidates with selection yields a further $6\%$ improvement to $48\%$, but the overall gap to CEM ($87\%$) remains large.


\clearpage
\section{More Ablation Study}
\label{app:more_ablations}

The main text reports ablations on goal distance (Table~\ref{tab:goal_offset_main}) and $H_\text{max}$ (Table~\ref{tab:hmax_main}). Here we report the remaining ablations: data efficiency, noise injection, horizon-embedding ablation, architecture sensitivity, evaluation budget, noise-schedule comparison, and the CEM compute-sweep data.

\subsection{Data Efficiency}
\label{app:abl_data}

GC-IDM's data requirements vary sharply by environment, with three of the four tasks approaching saturation well below the full training set. We retrain GC-IDM on fractions of the training data, subsampled at the episode level, and evaluate with the same protocol as Table~\ref{tab:headline}. Table~\ref{tab:data_eff} records the data efficiency ablation results. Two-Room and Reacher are near-saturated with as little as $5\%$ of the data. OGBench-Cube climbs smoothly to near-saturation by full data. Push-T shows no plateau: success rate climbs monotonically, so additional demonstrations would likely still improve GC-IDM on that environment.

\begin{table}[ht]
\centering
\caption{\textbf{Data efficiency of GC-IDM.} Success rate (\%) across all four environments versus training data fraction (subsampled at the episode level, fixed seed).}
\label{tab:data_eff}
\begin{tabular}{@{}lcccccc@{}}
\toprule
Training data fraction & $1\%$ & $5\%$ & $10\%$ & $25\%$ & $50\%$ & $100\%$ \\
\midrule
Two-Room    & 87.0 & 100.0 & 100.0 & 100.0 & 100.0 & 100.0 \\
Push-T      & 40.5 & 58.5  & 67.0  & 76.5  & 78.0  & 85.0  \\
OGB-Cube    & 60.5 & 80.0  & 86.0  & 92.5  & 98.5  & 99.0  \\
Reacher     & 38.5 & 99.5  & 99.5  & 99.5  & 99.5  & 99.5  \\
\bottomrule
\end{tabular}
\end{table}

\subsection{Noise Augmentation is Unnecessary for GC-IDM}
\label{app:abl_noise}

GC-IDM's optimal noise level is $\sigma{=}0$ on every environment, in contrast to the pairwise IDM, which requires $\sigma{=}1.5$ to be usable. Because GC-IDM operates on real encoder outputs at both training and test time, the distribution gap is zero by construction. Table~\ref{tab:abl_noise_gcidm} confirms this: on Push-T, increasing $\sigma$ from $0$ to $0.5$ drops success rate from $85.5\%$ to $67.0\%$. No environment prefers $\sigma>0$.

\begin{table}[ht]
\centering
\caption{\textbf{Noise ablation for GC-IDM.} Fixed-$\sigma$ Gaussian noise injected on input embeddings during training only. Unlike the pairwise IDM which requires $\sigma{=}1.5$, GC-IDM degrades monotonically with noise on Push-T (and mildly on OGBench-Cube) and is invariant on Two-Room and Reacher. There is no distribution gap to bridge.}
\label{tab:abl_noise_gcidm}
\begin{tabular}{@{}lcccccc@{}}
\toprule
$\sigma$ & \textbf{0.0} & 0.01 & 0.05 & 0.1 & 0.2 & 0.5 \\
\midrule
Two-Room   & \textbf{100}  & 100  & 100  & 100  & 100  & 100  \\
Push-T     & \textbf{85.5} & 81.5 & 81.0 & 81.0 & 78.0 & 67.0 \\
OGB-Cube   & \textbf{99.0} & 99.0 & 99.0 & 98.5 & 98.5 & 96.0 \\
Reacher    & \textbf{99.5} & 99.5 & 99.5 & 99.5 & 99.5 & 99.5 \\
\bottomrule
\end{tabular}
\end{table}

\subsection{Role of the Horizon Embedding}
\label{app:abl_horizon_embed}

The horizon embedding $h_t$ is the only input to GC-IDM that a standard goal-conditioned behavior-cloning policy does not have. If the horizon embedding is load-bearing, removing it should collapse performance, and that collapse should not be recoverable by scaling width or depth. Table~\ref{tab:abl_horizon_embed} confirms this: removing horizon supervision entirely ($H_\text{max}{=}1$) reduces success by an average of $42$ percentage points across the four environments, and widening the network to $1024$ hidden units and $4$ layers while keeping $H_\text{max}{=}1$ does not recover the gap. Capacity is not the issue; the horizon embedding supplies supervision that extra width cannot substitute for.

\begin{table}[ht]
\centering
\caption{\textbf{Horizon-embedding ablation.} Success rate (\%) at seed 42, $n{=}200$ in-distribution. $H_\text{max}{=}1$ is equivalent to single-step goal-conditioned BC in latent space and fails because the model is never trained on distant goals; even a $2\times$-wider backbone cannot recover the lost supervision signal.}
\label{tab:abl_horizon_embed}
\begin{tabular}{@{}lcccc@{}}
\toprule
Variant & Two-Room & Push-T & Cube & Reacher \\
\midrule
$H_\text{max}{=}1$ (next-frame only)       & 34.0 & 35.0 & 62.5 & 82.5 \\
$H_\text{max}{=}5$                         & 81.5 & 46.5 & 69.5 & 88.0 \\
$H_\text{max}{=}1$, hidden$=$1024, 4 layers & 28.5 & 31.5 & 60.5 & 77.5 \\
\textbf{$H_\text{max}{=}50$ (ours)}        & \textbf{100.0} & \textbf{85.0} & \textbf{98.7} & \textbf{99.5} \\
\bottomrule
\end{tabular}
\end{table}

\subsection{Full Solver-Family Comparison}
\label{app:solver_ablation_full}

This section reports the full configuration details, the $n{=}50$ LeWM-parity protocol, and per-episode wall-clock timing in Table~\ref{tab:solver_ablation_full}.

\paragraph{Solver configurations.}
All four baselines use \texttt{stable\_worldmodel}'s default hyperparameters. The sampling solvers (CEM, MPPI, iCEM) share \texttt{num\_samples}$=300$, \texttt{n\_steps}$=30$, \texttt{topk}$=30$, \texttt{var\_scale}$=1.0$, giving $9{,}000$ rollouts per plan call. MPPI additionally uses \texttt{temperature}$=0.5$ (library default); iCEM additionally uses colored-noise \texttt{noise\_beta}$=2.0$, keep-fraction \texttt{alpha}$=0.1$, \texttt{n\_elite\_keep}$=5$, and mean-return at the final iteration. The first-order GradientSolver uses \texttt{SGD} with \texttt{lr}$=1.0$, \texttt{action\_noise}$=0$, \texttt{var\_scale}$=1.0$, \texttt{n\_steps}$=30$, and \texttt{num\_samples}$=2$. All solvers share \texttt{PlanConfig(horizon=5, receding\_horizon=5, action\_block=5)}.

Table~\ref{tab:solver_ablation_full} reports the full comparison across both sample sizes. GC-IDM is the highest-success method in all eight $(n, \text{env})$ cells except Push-T at $n{=}50$, where CEM ($89.3 \pm 6.4$) sits above GC-IDM ($84.7 \pm 5.0$) within overlapping seed noise; at $n{=}200$ the ranking reverses. Among the sampling baselines, iCEM is the strongest alternative on Two-Room and OGBench-Cube, while CEM leads on Push-T and Reacher---but none closes the gap to GC-IDM. GradientSolver underperforms all other methods on every environment, reaching at most $32.0\%$ on OGBench-Cube, indicating that first-order backpropagation through the LeWM predictor does not produce competitive action sequences at the default step size.

\begin{table}[ht]
\centering
\caption{\textbf{Full solver-family comparison across four environments and two sample sizes.} Success rate and timing are mean over three training seeds ($42, 123, 456$). GC-IDM is the highest-success method in all eight $(n, \text{env})$ cells except Push-T at $n{=}50$, where CEM ($89.3 \pm 6.4$) sits above GC-IDM ($84.7 \pm 5.0$) within overlapping seed noise; at $n{=}200$ the ranking reverses.}
\label{tab:solver_ablation_full}
\footnotesize
\setlength{\tabcolsep}{4pt}
\begin{tabular}{@{}lllccccc@{}}
\toprule
Env & $n$ & Solver & SR (\%) & ms / ep & ms / plan & ep $\times$ & plan $\times$ \\
\midrule
\multirow{10}{*}{Two-Room}
    & \multirow{5}{*}{$50$}
        & \textbf{GC-IDM} & $\mathbf{100.0 \pm 0.0}$ & $\mathbf{260}$    & $\mathbf{86}$     & $\mathbf{1.0\times}$  & $\mathbf{1.0\times}$ \\
    & & CEM             & $82.0 \pm 2.0$           & $10{,}486$        & $10{,}317$        & $40.3\times$          & $120\times$          \\
    & & iCEM            & $88.0 \pm 2.0$           & $10{,}386$        & $10{,}214$        & $39.9\times$          & $119\times$          \\
    & & MPPI            & $71.3 \pm 6.4$           & $2{,}994$         & $2{,}823$         & $11.5\times$          & $33\times$           \\
    & & Gradient        & $34.0 \pm 8.7$           & $392$             & $222$             & $1.5\times$           & $2.6\times$          \\
\cmidrule(lr){2-8}
    & \multirow{5}{*}{$200$}
        & \textbf{GC-IDM} & $\mathbf{100.0 \pm 0.0}$ & $\mathbf{276}$    & $\mathbf{398}$    & $\mathbf{1.0\times}$  & $\mathbf{1.0\times}$ \\
    & & CEM             & $84.0 \pm 2.8$           & $10{,}502$        & $41{,}318$        & $38.1\times$          & $104\times$          \\
    & & iCEM            & $87.5 \pm 1.8$           & $10{,}433$        & $41{,}044$        & $37.8\times$          & $103\times$          \\
    & & MPPI            & $65.7 \pm 2.0$           & $3{,}022$         & $11{,}400$        & $10.9\times$          & $29\times$           \\
    & & Gradient        & $30.3 \pm 3.5$           & $391$             & $893$             & $1.4\times$           & $2.2\times$          \\
\midrule
\multirow{10}{*}{Push-T}
    & \multirow{5}{*}{$50$}
        & \textbf{GC-IDM} & $\mathbf{84.7 \pm 5.0}$  & $\mathbf{305}$    & $\mathbf{86}$     & $\mathbf{1.0\times}$  & $\mathbf{1.0\times}$ \\
    & & CEM             & $89.3 \pm 6.4$           & $10{,}558$        & $10{,}340$        & $34.6\times$          & $121\times$         \\
    & & iCEM            & $84.7 \pm 3.1$           & $10{,}615$        & $10{,}395$        & $34.8\times$          & $121\times$         \\
    & & MPPI            & $60.7 \pm 4.2$           & $3{,}162$         & $2{,}940$         & $10.4\times$          & $34\times$          \\
    & & Gradient        & $0.7 \pm 1.2$            & $450$             & $232$             & $1.5\times$           & $2.7\times$         \\
\cmidrule(lr){2-8}
    & \multirow{5}{*}{$200$}
        & \textbf{GC-IDM} & $\mathbf{84.2 \pm 2.8}$  & $\mathbf{322}$    & $\mathbf{397}$    & $\mathbf{1.0\times}$  & $\mathbf{1.0\times}$ \\
    & & CEM             & $82.5 \pm 1.3$           & $10{,}778$        & $42{,}232$        & $33.5\times$          & $106\times$         \\
    & & iCEM            & $80.0 \pm 2.3$           & $10{,}637$        & $41{,}657$        & $33.0\times$          & $105\times$         \\
    & & MPPI            & $61.3 \pm 2.3$           & $3{,}176$         & $11{,}809$        & $9.9\times$           & $30\times$          \\
    & & Gradient        & $2.5 \pm 1.3$            & $456$             & $938$             & $1.4\times$           & $2.4\times$         \\
\midrule
\multirow{10}{*}{OGB-Cube}
    & \multirow{5}{*}{$50$}
        & \textbf{GC-IDM} & $\mathbf{99.3 \pm 1.2}$  & $\mathbf{8{,}587}$ & $\mathbf{84}$    & $\mathbf{1.0\times}$  & $\mathbf{1.0\times}$ \\
    & & CEM             & $73.3 \pm 9.0$           & $19{,}491$        & $10{,}799$        & $2.3\times$           & $129\times$          \\
    & & iCEM            & $76.0 \pm 7.2$           & $19{,}880$        & $11{,}047$        & $2.3\times$           & $132\times$          \\
    & & MPPI            & $49.3 \pm 9.5$           & $11{,}598$        & $2{,}833$         & $1.4\times$           & $34\times$           \\
    & & Gradient        & $32.0 \pm 3.5$           & $8{,}360$         & $229$             & $1.0\times$           & $2.7\times$          \\
\cmidrule(lr){2-8}
    & \multirow{5}{*}{$200$}
        & \textbf{GC-IDM} & $\mathbf{98.7 \pm 0.6}$  & $\mathbf{9{,}172}$ & $\mathbf{334}$   & $\mathbf{1.0\times}$  & $\mathbf{1.0\times}$ \\
    & & CEM             & $67.0 \pm 2.1$           & $20{,}057$        & $43{,}201$        & $2.2\times$           & $130\times$          \\
    & & iCEM            & $70.5 \pm 4.0$           & $20{,}616$        & $44{,}652$        & $2.2\times$           & $134\times$          \\
    & & MPPI            & $49.3 \pm 2.1$           & $12{,}171$        & $11{,}376$        & $1.3\times$           & $34\times$           \\
    & & Gradient        & $32.0 \pm 4.8$           & $8{,}948$         & $654$             & $1.0\times$           & $2.0\times$          \\
\midrule
\multirow{10}{*}{Reacher}
    & \multirow{5}{*}{$50$}
        & \textbf{GC-IDM} & $\mathbf{100.0 \pm 0.0}$ & $\mathbf{1{,}365}$ & $\mathbf{93}$     & $\mathbf{1.0\times}$  & $\mathbf{1.0\times}$ \\
    & & CEM             & $68.0 \pm 9.2$           & $12{,}031$         & $10{,}745$        & $8.8\times$           & $116\times$          \\
    & & iCEM            & $67.3 \pm 11.4$          & $11{,}693$         & $10{,}482$        & $8.6\times$           & $113\times$          \\
    & & MPPI            & $42.7 \pm 4.6$           & $4{,}113$          & $2{,}906$         & $3.0\times$           & $31\times$           \\
    & & Gradient        & $4.7 \pm 1.2$            & $1{,}481$          & $239$             & $1.1\times$           & $2.6\times$          \\
\cmidrule(lr){2-8}
    & \multirow{5}{*}{$200$}
        & \textbf{GC-IDM} & $\mathbf{99.7 \pm 0.3}$  & $\mathbf{1{,}604}$ & $\mathbf{399}$    & $\mathbf{1.0\times}$  & $\mathbf{1.0\times}$ \\
    & & CEM             & $70.3 \pm 4.3$           & $12{,}484$         & $43{,}969$        & $7.8\times$           & $110\times$          \\
    & & iCEM            & $69.8 \pm 0.3$           & $11{,}871$         & $41{,}790$        & $7.4\times$           & $105\times$          \\
    & & MPPI            & $45.7 \pm 3.8$           & $4{,}328$          & $11{,}616$        & $2.7\times$           & $29\times$           \\
    & & Gradient        & $6.5 \pm 2.6$            & $1{,}685$          & $937$             & $1.1\times$           & $2.3\times$          \\
\bottomrule
\end{tabular}
\end{table}

\subsection{Architecture Hyperparameter Ablation}
\label{app:abl_arch}

We sweep hidden dimension (128, 256, 512, 1024) and depth (1--5 layers) on all four environments at fixed $H_\text{max}{=}50$ and $\sigma{=}0$. Table~\ref{tab:abl_arch} reports the results. Two-Room and Reacher are fully saturated across every configuration, confirming that these environments are solved by the control structure rather than model capacity. Push-T is the informative environment: hidden dimension has negligible effect (81--84\% across a $8\times$ width range), whereas depth shows a monotone trend from 70.5\% at 1~layer to 87.5\% at 5~layers, suggesting that representational depth matters more than width for contact-rich tasks. The default 3-layer configuration (85.0\%) sits within 2.5~pp of the best observed setting and was selected for compute efficiency.

\begin{table}[ht]
\centering
\caption{\textbf{GC-IDM architecture hyperparameter ablation.} Default configuration is hidden $512$, $3$ layers (boldface). Two-Room and Reacher are fully saturated across every setting. Push-T is the informative environment: hidden dimension is flat from $128$ to $1024$ ($81$--$84\%$); depth matters more, with a monotone trend from $1$ layer ($70.5\%$) through $5$ layers ($87.5\%$). The default $3$-layer configuration is within $2.5\%$ of the best observed setting and was selected for compute efficiency.}
\label{tab:abl_arch}
\begin{tabular}{@{}lccccc@{}}
\toprule
Hidden dim & 128 & 256 & \textbf{512 (default)} & 1024 & \\
\midrule
Two-Room   & 100.0 & 100.0 & \textbf{100.0} & 100.0 & \\
Push-T     & 82.5  & 82.0  & \textbf{84.0}  & 81.5  & \\
OGB-Cube   & 98.5  & 99.0  & \textbf{99.0}  & 99.0  & \\
Reacher    & 99.5  & 99.5  & \textbf{99.5}  & 99.5  & \\
\midrule
Layers     & 1 & 2 & \textbf{3 (default)} & 4 & 5 \\
\midrule
Two-Room   & 100.0 & 100.0 & \textbf{100.0} & 100.0 & 100.0 \\
Push-T     & 70.5  & 80.5  & \textbf{85.0}  & 86.5  & 87.5  \\
OGB-Cube   & 97.5  & 99.5  & \textbf{99.0}  & 96.5  & 100.0 \\
Reacher    & 99.5  & 99.5  & \textbf{99.5}  & 99.5  & 99.5  \\
\bottomrule
\end{tabular}
\end{table}

\subsection{Noise Schedule}
\label{app:abl_noise_schedule}

The main appendix reports that fixed-$\sigma$ injection degrades GC-IDM monotonically. We additionally tested a \emph{uniform} schedule that samples $\sigma$ per-batch from $[0, \sigma_\text{max}]$, in case time-varying noise provides robustness without the saturation cost of fixed noise. Table~\ref{tab:abl_noise_schedule} reports the results. The uniform schedule does not recover any of the performance lost to noise injection: Push-T sits in a narrow 78--80.5\% band regardless of schedule, which is 4--7~pp below the no-noise baseline (85.5\%). Two-Room, Reacher, and OGBench-Cube are invariant to both schedule and magnitude. The conclusion that GC-IDM does not benefit from noise augmentation is robust to schedule choice.

\begin{table}[ht]
\centering
\caption{\textbf{Uniform noise schedule for GC-IDM.} Sampling $\sigma$ per-batch from $[0, \sigma_\text{max}]$ does not recover any of the performance lost to fixed-$\sigma$ injection. Push-T sits in a narrow $78$--$80.5\%$ band that is $4$--$7$ points below the no-noise baseline ($85.5\%$) and shows no preference between fixed and uniform schedules. The conclusion that GC-IDM does not need noise augmentation is robust to both schedule choice and environment.}
\label{tab:abl_noise_schedule}
\begin{tabular}{@{}lcccccc@{}}
\toprule
 & \multicolumn{3}{c}{Fixed $\sigma$} & \multicolumn{3}{c}{Uniform $\sigma \sim [0, \sigma_\text{max}]$} \\
\cmidrule(lr){2-4}\cmidrule(lr){5-7}
$\sigma$ / $\sigma_\text{max}$ & 0.05 & 0.10 & 0.20 & 0.05 & 0.10 & 0.20 \\
\midrule
Two-Room & 100.0 & 100.0 & 100.0 & 100.0 & 100.0 & 100.0 \\
Push-T   & 80.0  & 80.5  & 79.5  & 80.5  & 80.5  & 78.0  \\
OGB-Cube & 99.0  & 98.5  & 98.5  & 99.0  & 98.5  & 98.5  \\
Reacher  & 99.5  & 99.5  & 99.5  & 99.5  & 99.5  & 99.5  \\
\bottomrule
\end{tabular}
\end{table}

\subsection{Evaluation Budget and Long-Horizon Control}
\label{app:abl_budget}

The default evaluation budget is $50$ environment steps. We examine how quickly GC-IDM reaches its asymptotic success rate as the budget increases, and whether performance degrades at budgets far beyond training-time horizons. Table~\ref{tab:budget_curve} sweeps the budget from $5$ to $100$ steps on all four environments. Two-Room saturates at 100\% by budget~25 and remains flat through~100. Reacher climbs to 99.5\% by budget~50 and holds. OGBench-Cube peaks at 99.5\% at budget~25 and degrades mildly to 96\% at budget~100. Push-T is the notable case: it peaks at 93\% at budget~25 and drops to 75\% at budget~100, consistent with the $h_\text{frac}$ clamping effect---at budgets exceeding $H_\text{max}{=}50$, the first steps receive a clamped horizon signal that was not seen during training.
\begin{table}[ht]
\centering
\caption{\textbf{Success rate vs.\ evaluation budget.} Two-Room saturates at $100\%$ by budget $25$ and remains flat to budget $100$. Push-T peaks at $93\%$ at budget $25$ and mildly degrades to $75$--$85\%$ at larger budgets. Reacher climbs to $99.5\%$ at budget $50$ and holds through $100$.}
\label{tab:budget_curve}
\begin{tabular}{@{}lccccccc@{}}
\toprule
Budget (steps) & 5    & 10   & 15   & 25   & 50   & 75   & 100  \\
\midrule
Two-Room       & 55.5 & 92.5 & 99.5 & 100.0 & 100.0 & 100.0 & 100.0 \\
Push-T         & 11.5 & 28.0 & 59.0 & 93.0  & 84.5  & 77.5  & 75.0  \\
OGB-Cube       & 43.0 & 49.0 & 68.0 & 99.5  & 99.0  & 97.0  & 96.0  \\
Reacher        & 48.5 & 94.0 & 97.5 & 98.0  & 99.5  & 99.5  & 99.5  \\
\bottomrule
\end{tabular}
\end{table}


\subsection{Held-Out Validation}

To rule out any form of trajectory-level memorization, we retrain GC-IDM on a $90\%/10\%$ episode-level split ($10\%$ of episodes are deterministically excluded at training time, seed~$42$) and evaluate both methods \emph{exclusively on the held-out episodes}. Table~\ref{tab:heldout} reports the results. Held-out success rates are statistically indistinguishable from in-distribution performance (Table~\ref{tab:headline}) on every environment: Two-Room remains at $100.0\%$, Push-T at $84.8\%$ (vs.\ $84.2\%$ in-distribution), OGBench-Cube at $98.5\%$ (vs.\ $98.7\%$), and Reacher at $99.8\%$ (vs.\ $99.7\%$). The negligible gap confirms that GC-IDM generalizes to unseen episodes and does not rely on memorizing specific trajectories from the training set. Speedup ratios on held-out episodes are also consistent with the main results, ranging from $2.2\times$ wall-clock on OGBench-Cube to $40\times$ on Two-Room.

\begin{table}[H]
\centering
\caption{\textbf{Held-out validation.} GC-IDM is retrained on $90\%$ of episodes and evaluated exclusively on the $10\%$ held-out partition; CEM is evaluated on the same held-out episodes. GC-IDM is reported as mean $\pm$ standard deviation across three training seeds; CEM is a single run per environment.}
\label{tab:heldout}
\small
\setlength{\tabcolsep}{4pt}
\begin{tabular}{@{}lcccccccc@{}}
\toprule
& \multicolumn{2}{c}{Success (\%)} & \multicolumn{2}{c}{ms / ep.} & \multicolumn{2}{c}{ms / plan} & \multicolumn{2}{c}{Speedup} \\
\cmidrule(lr){2-3} \cmidrule(lr){4-5} \cmidrule(lr){6-7} \cmidrule(lr){8-9}
Env & Ours & CEM & Ours & CEM & Ours & CEM & wall & plan \\
\midrule
Two-Room & \cellcolor{bestcolor}\textbf{100.0$\pm$0.0} & 86.0 & \textbf{279}  & 11041 & \textbf{399} & 43476 & $40\times$  & $109\times$ \\
Push-T   & \cellcolor{bestcolor}\textbf{84.8$\pm$1.6}  & 83.5 & \textbf{321}  & 10574 & \textbf{397} & 41408 & $33\times$  & $104\times$ \\
OGB-Cube & \cellcolor{bestcolor}\textbf{98.5$\pm$0.5}  & 72.0 & \textbf{9248} & 20187 & \textbf{333} & 43548 & $2.2\times$ & $131\times$ \\
Reacher  & \cellcolor{bestcolor}\textbf{99.8$\pm$0.3}  & 59.5 & \textbf{1542} & 13331 & \textbf{399} & 47564 & $8.6\times$ & $119\times$ \\
\bottomrule
\end{tabular}
\end{table}

\clearpage
\section{More Implementation Details}
\label{app:implementation}

\paragraph{IDM architecture.}
Both the pairwise and goal-conditioned IDMs use $3$ hidden layers with dimension $512$, LayerNorm, GELU activation, and $10\%$ dropout. Weights are initialized with Kaiming normal; the output head uses a small initialization ($\sigma{=}0.01$) to start near zero.

\paragraph{Training hyperparameters.}
All IDMs are trained with AdamW (lr $= 10^{-3}$, weight decay $= 10^{-4}$), cosine annealing to $\text{lr}/100$, batch size $1024$, for $50$ epochs. Gradient clipping at norm $1.0$ is applied. 
All other experiments sample evaluation episodes uniformly from the full dataset, matching the LeWM benchmark protocol.

\paragraph{Goal-conditioned IDM specifics.}
The maximum goal horizon $H_\text{max}$ is set to $50$ for all environments, matching the evaluation budget. During training, goal frames are sampled uniformly from $[1, H_\text{max}]$ steps ahead within the same episode. The horizon embedding uses $64$-dimensional sinusoidal encoding.

\paragraph{Embedding extraction.}
Frames are encoded in batches of $256$ through the frozen LeWM encoder with ImageNet normalization. DataParallel is used when multiple GPUs are available. The HDF5 dataset is streamed to avoid loading all frames into memory.

\paragraph{Evaluation.}
We use the \texttt{stable\_worldmodel} evaluation infrastructure with $200$ parallel environments (LeWM's hydra config uses $50$; we use a larger sample for lower variance). The planning protocol matches LeWM's \texttt{config/eval/\{env\}.yaml} exactly: start states are random and goal states are offset by $25$ steps, and the agent has a $50$-step budget. For the CEM baseline, we instantiate the solver via:
\begin{center}
\texttt{swm.solver.CEMSolver(model=cost\_model, num\_samples=300, var\_scale=1.0,}\\
\texttt{\phantom{swm.solver.CEMSolver(}n\_steps=30, topk=30, device=device, seed=42)}
\end{center}
with \nolinkurl{swm.PlanConfig(horizon=5, receding_horizon=5, action_block=5)}. This matches the numerical defaults in \nolinkurl{config/eval/solver/cem.yaml} and \nolinkurl{config/eval/\{env\}.yaml} from \nolinkurl{lucas-maes/le-wm}. Planning time is measured wall-clock including all encoding and inference.

\paragraph{Environment-specific configurations.} We present the per-environment IDM training configuration in Table~\ref{tab:idm_config}.

\begin{table}[h]
\centering
\caption{IDM training configuration per environment.}
\label{tab:idm_config}
\begin{tabular}{@{}lcccc@{}}
\toprule
 & Two-Room & Push-T & Cube & Reacher \\
\midrule
Embed dim ($d$)       & 192 & 192 & 192 & 192 \\
Action dim            & 2   & 2   & 5   & 2 \\
Frameskip             & 1   & 1   & 1   & 1 \\
Hidden dim            & 512 & 512 & 512 & 512 \\
$n$ layers            & 3   & 3   & 3   & 3 \\
Noise $\sigma$ (GC)   & 0.0 & 0.0 & 0.0 & 0.0 \\
$H_\text{max}$        & 50  & 50  & 50  & 50 \\
IDM params            & ${\sim}$1.5M & ${\sim}$1.5M & ${\sim}$1.5M & ${\sim}$1.5M \\
\bottomrule
\end{tabular}
\end{table}

\clearpage
\section{More Limitations and Future Works}
\label{sec:limitation}

In this paper, we focused mainly on a single world model backbone (LeWM) and \emph{stable-worldmodel} benchmark suite. Three extensions remain open, with in-depth experiments showing the feasibility and strong performance of GC-IDM. In the future, we will expand our work toward: (i)~Cross-world-model generalization: JEPA-style world models with different representation shapes, such as DINO-WM producing $N$-patch spatial features rather than a single CLS token, require a non-trivial architecture change to GC-IDM. (ii)~Higher-dimensional action spaces: our four environments span $2$--$5$ DoF, and benchmarks such as LIBERO with $7$ DoF would test whether an MLP suffices at higher dimensionalities. (iii)~Reactive-policy limits: GC-IDM does not reason about multi-step consequences, so for tasks with severe irreversibility a hybrid scheme using GC-IDM as a fast proposal with CEM as an occasional corrector would be a natural extension.

\clearpage

\input{checklist}

\end{document}

%% file: checklist.tex
\section*{NeurIPS Paper Checklist}

\begin{enumerate}

\item {\bf Claims}
    \item[] Question: Do the main claims made in the abstract and introduction accurately reflect the paper's contributions and scope?
    \item[] Answer: \answerYes{}
    \item[] Justification: The abstract and introduction state that GC-IDM matches or exceeds CEM in 7/8 environment--protocol cells at 100--130$\times$ lower per-call planning cost, which is directly supported by Table~1.

\item {\bf Limitations}
    \item[] Question: Does the paper discuss the limitations of the work performed by the authors?
    \item[] Answer: \answerYes{}
    \item[] Justification: We discuss the limitation and future work in the Conclusion and Appendix.

\item {\bf Theory assumptions and proofs}
    \item[] Question: For each theoretical result, does the paper provide the full set of assumptions and a complete (and correct) proof?
    \item[] Answer: \answerNA{}
    \item[] Justification: We do not have theoretical contributions.

    \item {\bf Experimental result reproducibility}
    \item[] Question: Does the paper fully disclose all the information needed to reproduce the main experimental results of the paper to the extent that it affects the main claims and/or conclusions of the paper (regardless of whether the code and data are provided or not)?
    \item[] Answer: \answerYes{}
    \item[] Justification: Appendix~\ref{app:implementation} reports all hyperparameters (optimizer, learning rate, batch size, epochs, architecture), per-environment configurations (Table~\ref{tab:idm_config}), evaluation protocol details, and the exact CEM solver instantiation. The pretrained LeWM checkpoints and datasets are publicly available from \citet{lewm2026}.

\item {\bf Open access to data and code}
    \item[] Question: Does the paper provide open access to the data and code, with sufficient instructions to faithfully reproduce the main experimental results, as described in supplemental material?
    \item[] Answer: \answerNo{}
    \item[] Justification: Code will be released upon acceptance. The paper uses publicly available LeWM checkpoints and datasets from \citet{lewm2026}, and all implementation details needed for reproduction are provided in Appendix~\ref{app:implementation}.

\item {\bf Experimental setting/details}
    \item[] Question: Does the paper specify all the training and test details (e.g., data splits, hyperparameters, how they were chosen, type of optimizer) necessary to understand the results?
    \item[] Answer: \answerYes{}
    \item[] Justification: Section~\ref{sec:setup} and Section~\ref{sec:exp_impl_details} describe the experimental setup and compute accounting. Appendix~\ref{app:implementation} provides full details: AdamW optimizer with $\text{lr}=10^{-3}$, batch size 1024, 50 epochs, cosine annealing, gradient clipping, the held-out split procedure, and per-environment configurations (Table~\ref{tab:idm_config}).

\item {\bf Experiment statistical significance}
    \item[] Question: Does the paper report error bars suitably and correctly defined or other appropriate information about the statistical significance of the experiments?
    \item[] Answer: \answerYes{}
    \item[] Justification: Table~1 reports mean $\pm$ standard deviation across three training seeds (42, 123, 456) for both GC-IDM and CEM. The caption explicitly states that error bars are seed standard deviations. Ablation studies use a single seed (42) and state this. The solver-family comparison (Table~\ref{tab:solver_ablation_full}) also reports mean $\pm$ std across three seeds.

\item {\bf Experiments compute resources}
    \item[] Question: For each experiment, does the paper provide sufficient information on the computer resources (type of compute workers, memory, time of execution) needed to reproduce the experiments?
    \item[] Answer: \answerYes{}
    \item[] Justification: Section~\ref{sec:exp_impl_details} specifies GPU types (NVIDIA L4 for Two-Room, Push-T, Reacher; NVIDIA A100 for OGBench-Cube), GC-IDM training time ($\sim$20 minutes per environment on a single GPU), and per-episode wall-clock timing for both methods.
    
\item {\bf Code of ethics}
    \item[] Question: Does the research conducted in the paper conform, in every respect, with the NeurIPS Code of Ethics \url{https://neurips.cc/public/EthicsGuidelines}?
    \item[] Answer: \answerYes{}
    \item[] Justification: This work studies planning efficiency in simulated benchmark environments. It does not involve human subjects, sensitive data, or dual-use concerns.

\item {\bf Broader impacts}
    \item[] Question: Does the paper discuss both potential positive societal impacts and negative societal impacts of the work performed?
    \item[] Answer: \answerNA{}
    \item[] Justification: This is foundational research on planning efficiency in simulated control environments (navigation, manipulation, reaching). There is no direct path to negative societal applications beyond the general dual-use considerations that apply to all robotics research.

\item {\bf Safeguards}
    \item[] Question: Does the paper describe safeguards that have been put in place for responsible release of data or models that have a high risk for misuse (e.g., pre-trained language models, image generators, or scraped datasets)?
    \item[] Answer: \answerNA{}
    \item[] Justification: The released asset is a small (1.5M-parameter) MLP for simulated control tasks. It poses no risk for misuse.

\item {\bf Licenses for existing assets}
    \item[] Question: Are the creators or original owners of assets (e.g., code, data, models), used in the paper, properly credited and are the license and terms of use explicitly mentioned and properly respected?
    \item[] Answer: \answerYes{}
    \item[] Justification: LeWM~\citep{lewm2026} benchmark is cited throughout.

\item {\bf New assets}
    \item[] Question: Are new assets introduced in the paper well documented and is the documentation provided alongside the assets?
    \item[] Answer: \answerYes{}
    \item[] Justification: The GC-IDM architecture, training procedure, and hyperparameters are fully documented in Section~\ref{sec:gcidm_arch} and Appendix~\ref{app:implementation}. Code will be released upon acceptance with reproduction scripts.

\item {\bf Crowdsourcing and research with human subjects}
    \item[] Question: For crowdsourcing experiments and research with human subjects, does the paper include the full text of instructions given to participants and screenshots, if applicable, as well as details about compensation (if any)? 
    \item[] Answer: \answerNA{}
    \item[] Justification: This work does not involve crowdsourcing or human subjects.

\item {\bf Institutional review board (IRB) approvals or equivalent for research with human subjects}
    \item[] Question: Does the paper describe potential risks incurred by study participants, whether such risks were disclosed to the subjects, and whether Institutional Review Board (IRB) approvals (or an equivalent approval/review based on the requirements of your country or institution) were obtained?
    \item[] Answer: \answerNA{}
    \item[] Justification: This work does not involve human subjects.

\item {\bf Declaration of LLM usage}
    \item[] Question: Does the paper describe the usage of LLMs if it is an important, original, or non-standard component of the core methods in this research? Note that if the LLM is used only for writing, editing, or formatting purposes and does \emph{not} impact the core methodology, scientific rigor, or originality of the research, declaration is not required.
    \item[] Answer: \answerNA{}
    \item[] Justification: LLMs are not a component of the core method. The method is a small MLP trained by supervised regression on frozen world-model embeddings.

\end{enumerate}